\documentclass[twoside,11pt, preprint]{article}

\usepackage{dmlr2e}

\usepackage{graphicx}
\usepackage{amsmath}
\usepackage{float}
\usepackage{minted}
\usepackage{array}
\usepackage{comment}
\usepackage{threeparttablex}
\usepackage{booktabs}
\usepackage[utf8]{inputenc} % allow utf-8 input
\usepackage[T1]{fontenc}    % use 8-bit T1 fonts
\usepackage{xltabular}
\usepackage{afterpage}
\usepackage{pifont}

\newcommand{\cmark}{\ding{51}}%
\newcommand{\xmark}{\ding{55}}%

\usepackage{lastpage}
\dmlrheading{23}{2025}{1-\pageref{LastPage}}{9/23; Revised 10/23}{11/23}{21-0000}{Simson et al.} % Insert the dates and author names. This is not relevant if it is yet being reviewed, i.e., \documentclass[twoside,11pt, preprint]{article}
\ShortHeadings{Bias Begins with Data}{FairGround}
\firstpageno{1}

\begin{document}

\title{Bias Begins with Data:\\
The \textit{FairGround} Corpus for Robust and Reproducible Research on Algorithmic Fairness}

\author{\name Jan Simson \email jan.simson@lmu.de \\
    \addr Department of Statistics, LMU Munich\\
    Munich Center for Machine Learning (MCML)\\
    Munich, 80539, Germany
\AND
    \name Alessandro Fabris \email alessandro.fabris@units.it \\
    \addr 
    University of Trieste\\
    Trieste, 33170, Italy
\AND
    \name Cosima Fröhner \email c.froehner@campus.lmu.de\\
    \addr Department of Statistics, LMU Munich\\
    Munich, 80539, Germany
\AND
    \name Frauke Kreuter \email frauke.kreuter@lmu.de \\
    \addr Department of Statistics, LMU Munich\\
    Munich Center for Machine Learning (MCML)\\
    Munich, 80539, Germany
\AND
    \name Christoph Kern \email christoph.kern@lmu.de \\
    \addr Department of Statistics, LMU Munich\\
    Munich Center for Machine Learning (MCML)\\
    Munich, 80539, Germany
}

\editor{My editor}

\maketitle

\begin{abstract}%   <- trailing '%' for backward compatibility of .sty file

As machine learning (ML) systems are increasingly adopted in high-stakes decision-making domains, ensuring fairness in their outputs has become a central challenge. At the core of fair ML research are the datasets used to investigate bias and develop mitigation strategies. Yet, much of the existing work relies on a narrow selection of datasets--often arbitrarily chosen, inconsistently processed, and lacking in diversity--undermining the generalizability and reproducibility of results.

To address these limitations, we present \textit{FairGround}: a unified framework, data corpus, and Python package aimed at advancing reproducible research and critical data studies in fair ML classification. 
FairGround currently comprises 44 tabular datasets, each annotated with rich fairness-relevant metadata. Our accompanying Python package standardizes dataset loading, preprocessing, transformation, and splitting, streamlining experimental workflows. By providing a diverse and well-documented dataset corpus along with robust tooling, FairGround enables the development of fairer, more reliable, and more reproducible ML models. All resources are publicly available to support open and collaborative research.

\end{abstract}


\section{Introduction}

The field of algorithmic fairness has grown rapidly, reflecting the increasing recognition of fairness as a core concern in machine learning \citep{mehrabi2021survey,pessach2023algorithmic}. 
Progress in this field is inevitably tied to data as the central ingredient to developing, testing and benchmarking more equitable algorithms. Given that these algorithms and fairness-enhancing techniques are often deployed in high-risk contexts [e.g., healthcare \citep{obermeyer2019racial, barda2020developing}, criminal justice \citep{angwin2016machine, carton2016police}, jobseeker profiling \citep{kern2024profiling, achterhold2025fairness}], systematic and transparent evaluations based on principled rather than ad-hoc selections of datasets are critical to understand which method works reliably under which conditions and which might not yet be ready for deployment.

Progress in Fair ML is challenged by (1) opacity in data practices and (2) critical limitations of the most prominent datasets currently used. 
A number of studies have shown that seemingly minor data processing and algorithmic design choices can significantly impact fairness outcomes, raising important questions about the robustness and generalizability of existing fairness interventions \citep{simson2024one,friedler2019comparative,caton2022impact}. Compounding these concerns, recent work has also highlighted reproducibility challenges that hinder consistent evaluation across settings \citep{cooper2024arbitrariness,simson2024lazy}. 
Furthermore, large-scale comparisons of fairness algorithms not only show strong sensitivity to data processing decisions, but also considerable performance differences between datasets, underlining the importance of the exact collection of data used for benchmarking and evaluation \citep{agrawal2021debiasing}.   
Unfortunately, current studies commonly still focus on a narrow set of benchmark datasets--such as \textit{Adult} \citep{kohavi1996scaling}, \textit{COMPAS} \citep{angwin2016machine} and \textit{German Credit} \citep{hofmann1994statlog}--which suffer from known limitations, including contrived prediction tasks, noisy data, and severe coding mistakes \citep{ding2021retiring,bao2022it,groemping2019south}. 
% folktables?
Taken together, these practices can lead to evaluations of fairness algorithms that are driven by methodological artifacts rather than representing reliable performance tests that justify the (non-)deployment of a given method in practice. 
% unclear licensing?

\begin{figure}
    \centering
    \includegraphics[width=0.8\linewidth]{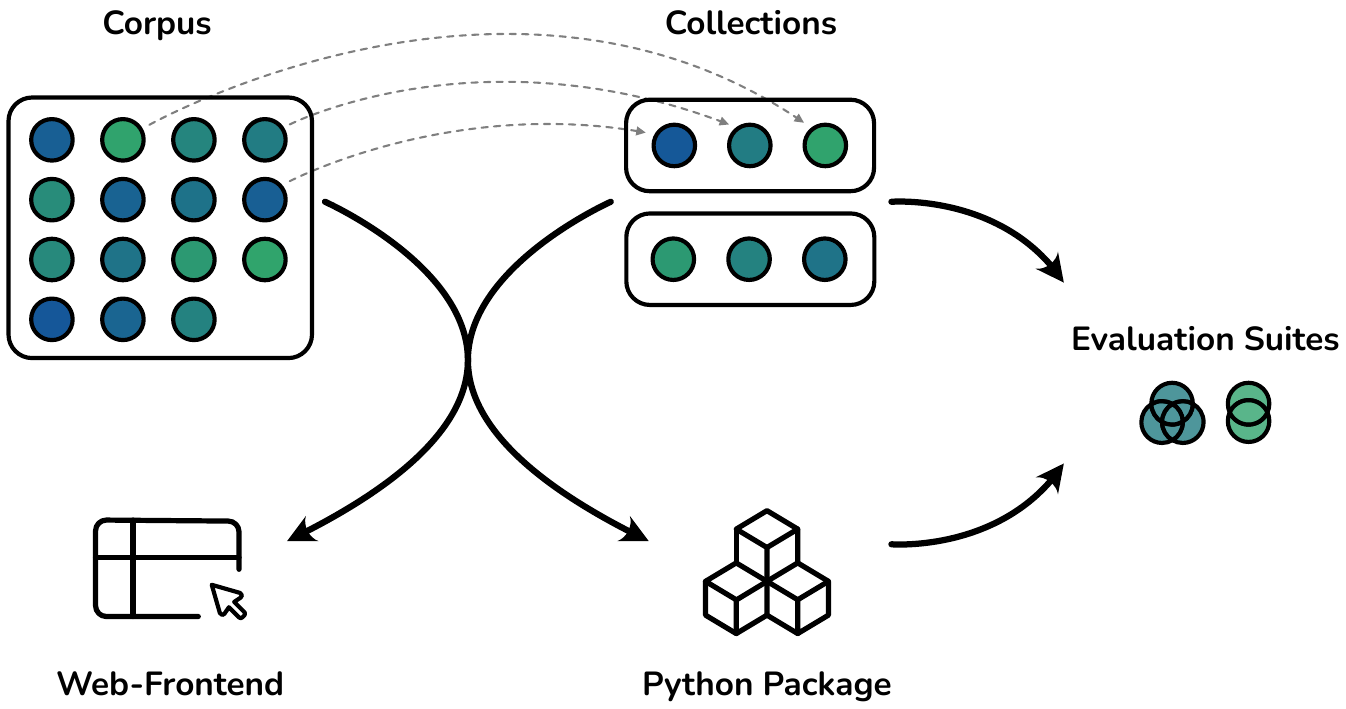}
    \caption{\textbf{The different components in the FairGround corpus.} We provide a comprehensive corpus of datasets and extract diverse collections of datasets via a selection algorithm. Both the corpus, collections and individual datasets are made accessible via a Python package and web-frontend. Collections paired with reproducible dataset loading and preparation allows for novel evaluation suites.}
    \label{fig:framework}
\end{figure}

Addressing these limitations, this work introduces \textit{FairGround}: a framework that emphasizes reproducible data processing pipelines, standardized evaluation protocols, and diverse collections of datasets tailored to specific needs (Figure \ref{fig:framework}). 
Our corpus contains 136 scenarios, i.e. combinations of 44 tabular datasets with available sensitive attributes. Each dataset %in our corpus of 44 tabular datasets
comes with rich metadata (35 annotated and 27 computed meta-features), which allows for a principled selection of benchmarking collections and for failure testing of algorithms to identify data scenarios under which a proposed method struggles to perform. We further provide a data selection algorithm and associated collections of datasets that are small but diverse, i.e., present challenging scenarios with data that capture the variability present in the larger corpus. Our Python package facilitates transparent data practices in fair ML through reproducible and standardized, but customizable, processing pipelines. 
With FairGround, we contribute infrastructure that supports more robust and generalizable evaluation of fairness-aware machine learning methods.

\section{Related work}

\subsection{Comparing fairness-enhancing algorithms}

A number of prior studies have carried out systematic comparisons of fairness-enhancing algorithms across different settings \citep{friedler2019comparative, agrawal2021debiasing, biswas2020machine, cruz2024unprocessinga, defrance2024abcfair, han2023ffb, hort2021fairea, islam2021can, l.cardoso2019framework}. While these comparative efforts have contributed valuable insights, they are often constrained by a narrow and inconsistently chosen set of benchmark datasets. In many cases, dataset selection is neither well-documented nor critically examined, resulting in evaluations that are difficult to reproduce and limited in scope.

The broader field continues to face fundamental challenges related to reproducibility and transparency in experimental design \citep{simson2024lazy, cooper2024arbitrariness}. One prominent issue is the lack of principled approaches to dataset processing and selection. Many existing works make ad hoc or arbitrary choices when selecting datasets \citep{ding2021retiring, bao2022it, groemping2019south}, often relying on convenience or popularity rather than representativeness or relevance. These decisions can unintentionally bias results and restrict the generalizability of conclusions.
A core concern here is that the datasets typically used in fairness evaluations do not adequately reflect the diversity and complexity of real-world deployment scenarios. The dominance of a small set of benchmark datasets has led to evaluations that cover only a limited subset of the problem space fairness algorithms are meant to address \citep{fabris2022algorithmic}.

Compounding this, there remains little clarity around the specific data conditions under which fairness methods are expected to succeed or fail. Without a systematic understanding of these contexts, practitioners are left with limited guidance on which algorithms to apply in practice \citep{richardson2021towards,holstein2019improving}, reducing the effectiveness and reliability of fairness interventions in real-world systems.

\subsection{Fairness toolkits and data studies}

Fairness datasets have been examined from both granular and comparative perspectives. Some works offer deep, dataset-specific critiques \citep{bandy2021addressing,ding2021retiring,bao2022it,birhane2023into}, while others survey broader patterns across multiple datasets \citep{crawford2021excavating,fabbrizzi2022survey,fabris2022algorithmic,zhao2024taxonomy}. 
In parallel, fairness-focused toolkits such as AIF360 \citep{bellamy2018aif360}, Fairlearn \citep{weerts2023fairlearn}, and Aequitas \citep{jesus2024aequitas} implement popular algorithmic interventions and metrics, providing an accessible entry point for numerical comparisons—while including only a few illustrative datasets (Table \ref{tab:tab-quant-comparison}).
Despite overlapping goals, these two strands have remained largely disconnected. Toolkits often treat datasets as ancillary components and dataset-focused studies fail to produce machine-readable resources designed for seamless integration with software frameworks. Bridging critical data studies and fairness toolkits is essential for advancing the field, as meaningful integration can enable more rigorous, interpretable, and reproducible fairness evaluations--particularly by linking dataset properties to the behavior and impact of fairness interventions \citep{li2022data,favier2023how}. 
%Broader infrastructure efforts like OpenML \citep{OpenML2013} improve general ML reproducibility but do not address the specific requirements of fairness research, such as transparent preprocessing or group-level annotations.

A recent article, most closely related to ours, lists several fairness resources and provides a tool for data fetching, but does not address integrated processing or annotation pipelines \citep{hirzel2023suite}. In this paper, we address this gap by introducing a benchmark suite that combines (1) a curated corpus of datasets accompanied by rich quantitative and qualitative annotations, (2) reproducible data fetching and processing pipelines, and (3) standardized collections and evaluation protocols. Our annotations provide a foundation for aligning datasets with fairness-aware methods in a consistent, reproducible, and extensible manner.

\section{Framework}\label{sec:framework}

We introduce a unified framework of resources designed to support reproducible research and critical data studies in fair ML. While our current implementation focuses on tabular classification, which is prominent in fair ML research \cite{mehrabi2021survey,caton2024fairness}, the underlying design is broadly applicable to other contexts.

\subsection{Corpus}
\label{sec:corpus}

Building on and extending beyond prior surveys of datasets in fair ML research \citep{fabris2022algorithmic,le2022survey}, we compile a curated corpus of $N = 44$ tabular datasets. Each dataset is annotated with extensive fairness-relevant metadata, both quantitative and qualitative. While an additional 11 datasets were partially annotated, we excluded them from the final release due to issues such as dubious provenance or access restrictions (details in Section~\ref{sec:excluded-datasets}).

The corpus spans a wide range of dataset sizes, from 118 to over 3.2 million records, and 4 to 1,941 features. Most datasets originate from domains such as economics and law (each 23.4\%), followed by finance (12.7\%) and education (10.7\%). Geographical representation is notably skewed: nearly 60\% of datasets originate from the United States, with limited coverage from other regions (see Tables \ref{tab:datasets} and \ref{tab:country-dataset-distribution} for details).
The dataset metadata can be explored interactively at: \url{https://reliable-ai.github.io/fairground/}.

Following prior work \citep{fabris2022algorithmic,le2022survey}, we annotate each dataset with contextual information (e.g., dataset name, domain), data-specific attributes (e.g., geography, time period), and technical metadata required for loading and preparing the data. Where multiple variants of a dataset exist, each version is treated as a distinct entry (cf. Section~\ref{sec:annotation-procedure}). We also provide annotations relevant to fair ML tasks, including sensitive attribute selection, target variable definitions, and required preprocessing. While we do not claim our annotations are definitive, they serve as principled defaults that make implicit modeling decisions explicit, encouraging transparency in fair ML research \citep{simson2024lazy}. Full details on our annotation procedure are provided in Sections~\ref{sec:annotation-procedure} and \ref{sec:annotation-columns-appendix}.

In addition to manual annotations, we compute a range of metadata to support dataset selection, benchmarking, and critical analysis. This includes structural properties (e.g., missing values, feature types), statistical characteristics (e.g., bivariate correlations, sensitive AUC), and fairness-related properties (e.g., protected group prevalence, base rates, Gini-Simpson index) \citep{brzezinski2024properties,mecati2022detecting,holland2020dataset}. These computed metadata features are detailed in Appendix~\ref{sec:computed-meta-data} and integrated into our Python tooling for streamlined access.

\subsection{Infrastructure}
\label{sec:infra}

\begin{figure}
    \centering
    \includegraphics[width=1.0\linewidth]{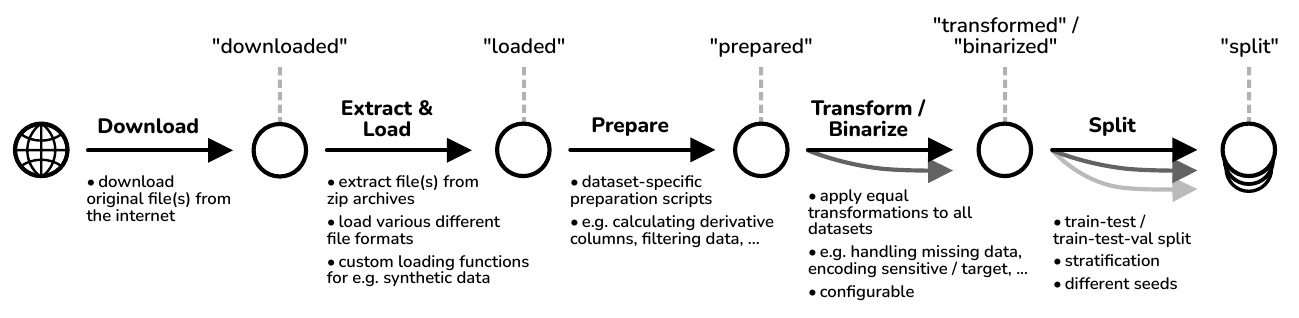}
    \caption{\textbf{The pipeline of steps involved when loading and processing a dataset in the package.} Datasets can be accessed / exported after each of the steps in the pipeline and most steps allow for configuration.}
    \label{fig:pipeline}
\end{figure}

% \subsubsection{Package}
\label{sec:package}

To enable reproducible and scalable use of the corpus, we provide a Python package that operationalizes our framework. This package automates dataset acquisition, preprocessing, transformation, and splitting, applying the annotations to prepare datasets for downstream fair ML tasks (Figure~\ref{fig:pipeline}).
The package supports diverse data formats and includes default transformations, such as standard feature selection, handling of missing values and encoding of sensitive attributes. We re-emphasize that defaults are not intended as universally correct, but rather as transparent baselines that can be fully customized. By surfacing and standardizing preprocessing decisions, the package encourages methodological rigor and reduces hidden variability in experimental pipelines \citep{simson2024one}.

In particular, FairGround supports the following transformations to export data in a readily usable format. Users can choose to retain either the complete set of columns in a dataset or only the essential subset, which includes frequently-used features, sensitive attributes, and the target variable (default). To handle missing values, the framework supports three options: dropping the entire column, removing only rows with missing values, or imputing missing values using the median (default for numerical) or a placeholder value (default for categorical). The target variable can be binarized in several ways: based on an annotated preferable label, redefined to reflect a majority/minority split, or automatically selected between these options depending on metadata availability (default is based on the preferable label if provided). When multiple sensitive attributes exist, users can keep them separate (default) or combine them into a single binary attribute that captures their intersection (default in the binarized setting). Sensitive attribute values can be left unchanged or grouped into majority and minority categories (again, grouping is the default in binarized datasets). For categorical features, FairGround supports either leaving them as-is or converting them into binary indicators via dummy encoding (default). To control for high cardinality in categorical or text fields, the package applies an optional limit--by default, restricting each categorical or text column to a maximum of 200 unique values, with less frequent categories grouped together once this limit is exceeded.

The package also supports automatic metadata extraction (see Section~\ref{sec:corpus}). Importantly, we avoid redistributing raw data directly to respect licensing constraints and datasets are instead downloaded from their original sources and optionally cached locally.

\noindent The package is open source and available at: \url{https://github.com/reliable-ai/fairground}

\noindent Releases are archived on Zenodo: \url{https://doi.org/10.5281/zenodo.17288596}

\noindent Package installation: \texttt{pip install fairml-datasets}

\noindent Package documentation: \url{https://reliable-ai.github.io/fairground/docs/}

\noindent Code examples are provided in Appendix~\ref{sec:package-examples}.

In parallel, we release an interactive website that allows browsing the dataset corpus, metadata, and example usage. The site also offers sample code for specific datasets and is available at \url{https://reliable-ai.github.io/fairground/}.

\subsection{Collections}

To further support reproducible benchmarking and targeted experimentation, we define several curated dataset collections derived from the full corpus with an extensible algorithm. These include: two collections (small and large) optimized for diversity in algorithmic performance; three collections with permissive licenses; and three collections emphasizing geographic diversity (Tables \ref{tab:de-correlated-datasets-large}--\ref{tab:geographically-diverse-datasets-all}).

Combined with standardized data splits from our package, which are critical to fair ML reproducibility \citep{friedler2019comparative}, these collections provide ready-to-use evaluation suites for fair ML development.

\section{Experiments}
\label{sec:experiments}

Leveraging the full FairGround dataset corpus, we conduct a series of experiments to systematically investigate the extent to which the choice of dataset influences the evaluation and observed performance of fairness-aware machine learning methods.

To reflect common practice in fairness research and enable broad coverage of methodological approaches, we evaluate a representative set of fairness-aware debiasing techniques spanning the three main intervention stages in the ML pipeline: \textit{pre-processing}, \textit{in-processing}, and \textit{post-processing}. Specifically, we compare the following seven algorithms: \textit{Learning Fair Representations} (pre) \citep{zemel2013learning}, \textit{Disparate Impact Remover} (pre) \citep{feldman2015certifying}, \textit{Adversarial Debiasing} (in) \citep{zhang2018mitigating}, \textit{Meta-Algorithm} (in) \citep{celis2019classification}, \textit{Rich Subgroup Fairness / GerryFair} (in) \citep{kearns2018preventing}, \textit{Grid Search Reduction} (in) \citep{agarwal2018reductions}, and \textit{Group-Specific Thresholds} (post) \citep{hardt2016equality}. We use logistic regression as a standard model for pre- and post-processing.

To satisfy the input constraints of all methods, datasets were converted to binarized numerical representations using the default transformation settings provided by our accompanying Python package (see Section~\ref{sec:package}). This ensures compatibility while preserving consistency across experiments.

Given that most fairness techniques are designed to optimize fairness with respect to a single sensitive attribute, we adopt a principled approach to define sensitive attribute configurations. For datasets containing fewer than four sensitive attributes, we evaluate all individual attributes and their pairwise intersections. For datasets with four or more sensitive attributes, we restrict evaluation to individual attributes to avoid combinatorial complexity. We refer to each combination of a dataset and its corresponding sensitive attribute selection as a unique \emph{\textbf{scenario}}.

We apply each of the seven processing methods and a baseline to each of the $n = 136$ datasets and sensitive attribute combinations (scenarios) across five separate seeds and train-test splits. This results in a total of $N = 5440$ different models that are trained and compared. For each model, we compute two commonly used measures of performance (\textit{Balanced Accuracy}, Eq.~\ref{eq:bacc}; \textit{F1 Score}, Eq.~\ref{eq:f1}) and two measures of algorithmic fairness (\textit{Equalized Odds Difference}, Eq.~\ref{eq:eod}; \textit{Demographic Parity Difference}, Eq.~\ref{eq:dpd}). The computational infrastructure (Section~\ref{sec:exp-infrastructure}) and software (Section~\ref{sec:software}) used for experiments are described in the technical appendix.

\subsection{Results}

\begin{figure}
    \centering
    \includegraphics[width=0.75\linewidth]{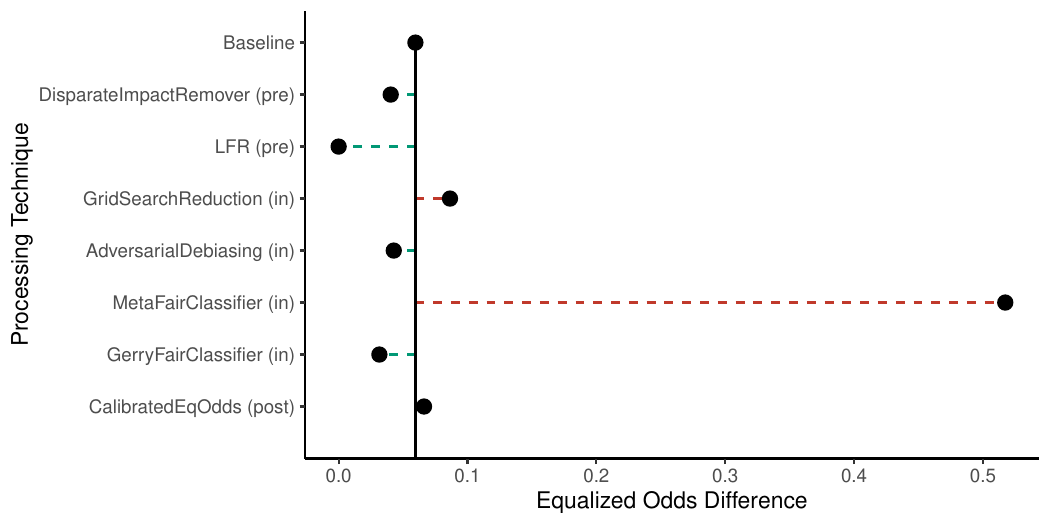}
    \caption{Scores from a single dataset (\textit{Bank}), scenario (sensitive attribute: \textit{Age}), seed (\textit{80539}), and metric (\textit{Equalized Odds Difference}) illustrating how delta scores with respect to baseline logistic regression are calculated. Delta scores correspond to dashed lines.}
    \label{fig-illustration-deltas-1.pdf}
\end{figure}

The experiments reveal substantial variation in both fairness and performance metrics across datasets and methods. F1 score, equalized odds difference, and demographic parity difference span the full $[0, 1]$ range, while balanced accuracy varies from approximately $0.2$ to $1.0$.
To facilitate comparisons, we compute delta scores—metric differences relative to a logistic regression baseline without fairness interventions (Eq.~\ref{eq:delta_score}). Figure~\ref{fig-illustration-deltas-1.pdf} illustrates this calculation for one dataset, scenario, seed, and metric, with dashed lines indicating differences from the baseline.

The overall distribution of delta scores across all four metrics is shown in Figure~\ref{fig:fig-hist-deltas-overview-1}. Importantly, fairness interventions often lead to minor deviations in scores, as highlighted by the large gray bar indicating an absolute change of $\leq$ 0.01, which correspond to scenarios where popular fair ML methods are ineffective. 
A sizable portion produces meaningful differences, typically reflecting the well-known tradeoff between fairness and performance \citep{menon2018cost,islam2021can}: improvements in fairness often coincide with declines in predictive accuracy.

\subsection{Rankings of Debiasing Techniques are not Stable}

\begin{figure}
    \centering
    \includegraphics[width=1.0\linewidth]{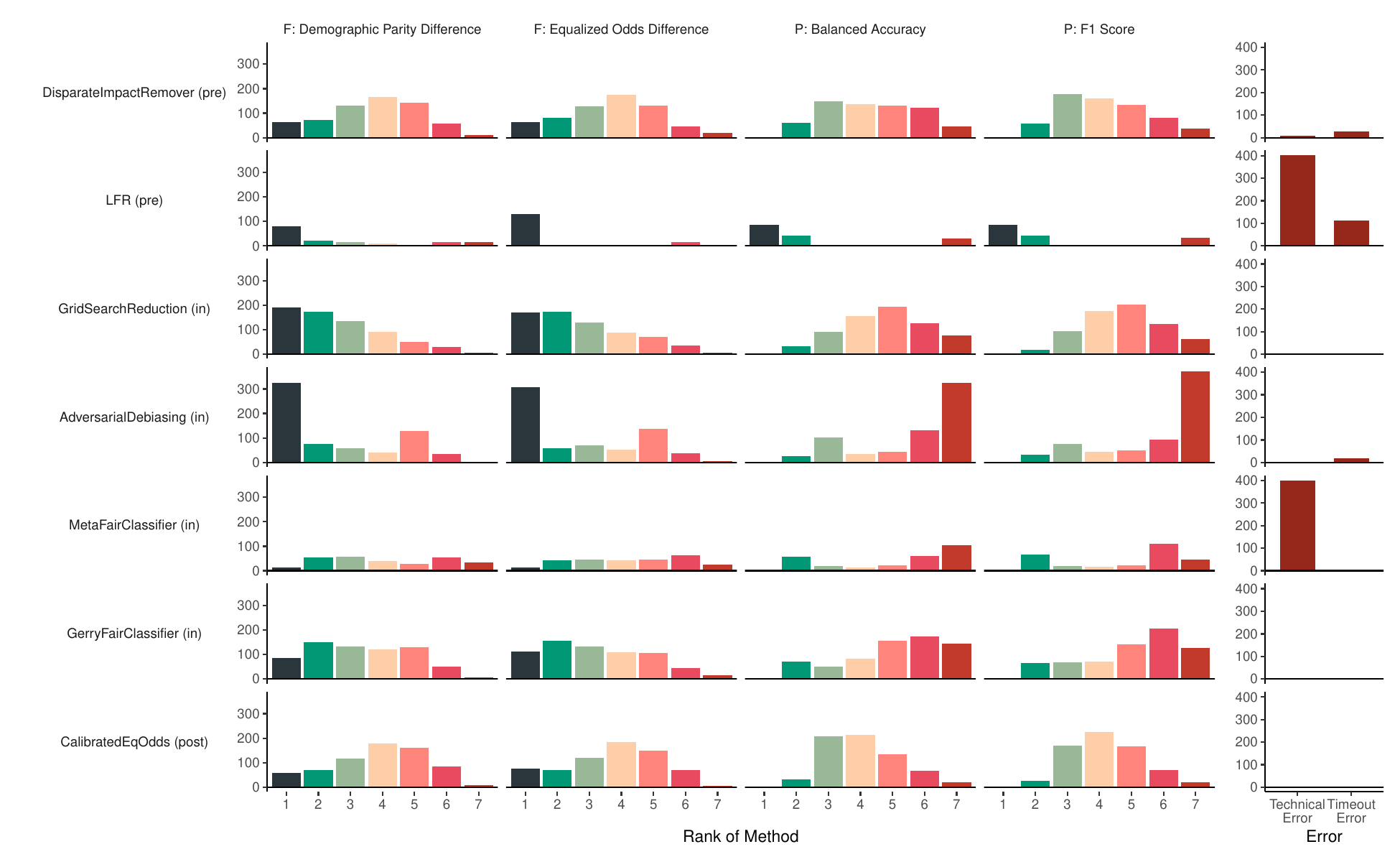}
    \caption{\textbf{Relative performance and efficacy of different fairness interventions is highly variable.} Relative ranking of different processing techniques across datasets and seeds (A), as well as prevalence of practical and timeout errors (B).}
    \label{fig:fig-ranks-combined-1}
\end{figure}

To reflect how practitioners might compare processing techniques in practice, we analyze the relative rankings of different methods. While some methods--such as \textit{LFR}, \textit{Grid Search Reduction}, and \textit{Adversarial Debiasing}--tend to rank favorably, their positions vary considerably across scenarios, and no single method consistently outperforms the rest (Figure~\ref{fig:fig-ranks-combined-1}).
High-performing methods often come with caveats. For instance, \textit{LFR} occasionally fails due to convergence issues or label collapse during rebalancing, rendering it unusable in some cases. \textit{Adversarial Debiasing} often presents sharp tradeoffs between fairness and predictive performance.
These variations are influenced by the dataset and scenario characteristics.

\subsection{Identifying Important Dataset Characteristics}
\label{sec:predicting}

\begin{figure}
    \centering
    \includegraphics[width=0.8\linewidth]{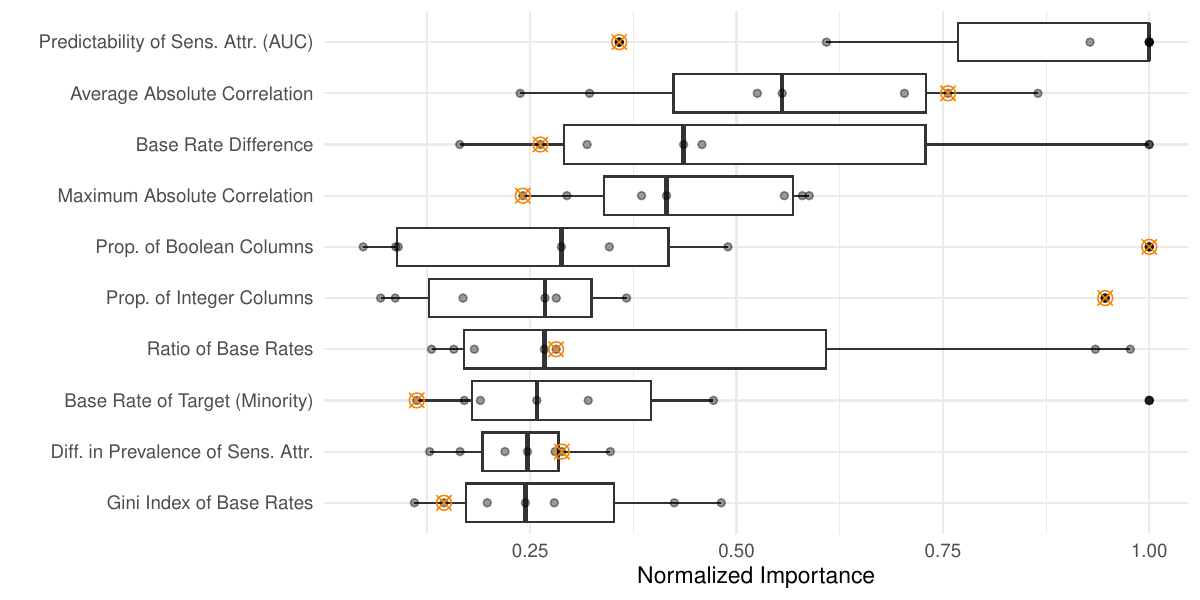}
    \caption{\textbf{The importance of different dataset characteristics can be highly variable between debiasing algorithms.} Normalized feature importance of the 10 most important computed metadata features to predict the difference in \textit{Equalized Odds Difference} across all processing methods, ordered by average importance. Feature importance for \textit{Adversarial Debiasing (in)} is highlighted in orange.}
    \label{fig:fig-pred-varimp-eqodds-1}
\end{figure}

To uncover which dataset properties affect method performance, we train simple machine learning models (random forests \citep{ho1995random}) for each debiasing technique. These models use only computed metadata (Sections~\ref{sec:corpus},~\ref{sec:computed-meta-data}) to predict method effectiveness across individual scenarios. As shown in Figure~\ref{fig:fig-pred-vs-observed-combined-1}, they capture substantial variance in observed outcomes.
We analyze feature importance scores from these models to assess which dataset characteristics matter most. Figure~\ref{fig:fig-pred-varimp-eqodds-1} displays importances for predicting \textit{Equalized Odds Difference}. A key trend is that the predictability of sensitive features from non-sensitive ones (\texttt{meta\_sens\_predictability\_roc\_auc}, top row) is influential across all methods. Base rate differences are critical for some techniques but negligible for others. These metadata-derived features help characterize the conditions under which fairness interventions are likely to succeed. Notably, \textit{Adversarial Debiasing}, highlighted in orange, relies less on sensitive attribute predictability and more on structural features such as the proportion of boolean and integer columns. Relative importances for other metrics appear in Figure~\ref{fig:fig-pred-varimp-combined-1}.

\subsection{Developing Diverse Collections of Datasets}
\label{sec:results-collections}

Evaluating fairness interventions across all possible datasets and scenarios is ideal but rarely feasible due to practical constraints like limited compute. To address this, we construct eight curated dataset collections, each optimized for a specific purpose.
We use a principled algorithm to construct subsets of scenarios that exhibit diverse properties. 
We explicitly target predictive accuracy and fairness properties by building a collection of scenarios whose pairwise spearman correlations of delta scores (Eq.~\ref{eq:delta_score}), across \textit{Balanced Accuracy} (Eq.~\ref{eq:bacc}), \textit{F1 Score} (Eq.~\ref{eq:f1}), \textit{Equalized Odds Difference} (Eq.~\ref{eq:eod}) and \textit{Demographic Parity Difference} (Eq.~\ref{eq:dpd}) are as low as possible. The underlying assumption is that datasets where debiasing techniques yield divergent fairness-performance tradeoffs make for more informative and challenging benchmarks.
The algorithm greedily builds collections by adding the least correlated scenario while fulfilling optional secondary constraints, including selecting only a single scenario per dataset. The algorithm supports two different cutoff values, providing either a fixed number of $k$ scenarios or a fixed upper bound for dataset correlation ($\bar{r}_{j\mathcal{C}} < \tau$) when added to the collection. The algorithm is described in detail in Section~\ref{sec:selection-algorithm}.
We use this selection process both to construct benchmark collections and to define default scenarios per dataset.
We demonstrate how the \textit{FairGround} corpus as well as its collections exhibit higher diversity in algorithm performance compared to other dataset collections (Table~\ref{tab:tab-copmarison-method-performance}).

\begin{figure}
    \centering
    \includegraphics[width=0.8\linewidth]{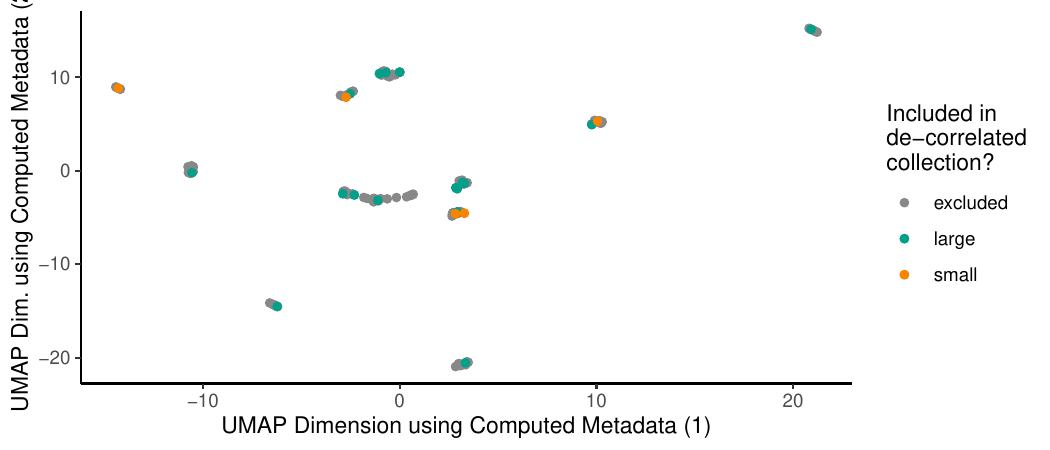}
    \caption{\textbf{Datasets in the de-correlated collections capture variability in the computed metadata features well.} Two dimensional mapping of datasets using UMAP on computed metadata features. Scenarios in the de-correlated collections are highlighted in different colors.}
    \label{fig:fig-meta-umap-benchmark-1}
\end{figure}

\textbf{De-Correlated Datasets}\quad We construct two benchmark collections using the correlation-based algorithm with cutoffs $k = 5$ and $\tau < 0$, yielding sets of $n = 5$ and $n = 22$ scenarios, respectively (Table~\ref{tab:de-correlated-datasets-large}). A UMAP projection \citep{mcinnes2018umap} from the high-dimensional space of computed metadata (Figure~\ref{fig:fig-meta-umap-benchmark-1}) confirms that selected datasets span a wide range of characteristics.

\textbf{Permissively Licensed Datasets}\quad To facilitate open sharing and reuse, we build three collections containing only datasets with permissive licenses. We construct these collections by only allowing datasets to be added to the collection which (1) have licensing information available and (2) are permissively licensed (e.g. Creative Commons, Apache, GNU licenses). One collection uses a fixed $k = 5$ cutoff, one uses a $\tau < 0$ threshold ($n = 16$), and one includes all permissively licensed datasets without filtering ($n = 32$). All three are listed in Table~\ref{tab:permissively-licensed-datasets-all}. We release these datasets in both prepared and binarized formats.

\textbf{Geographically Diverse Datasets}\quad To address regional bias, we create three collections ensuring that no two datasets originate from the same country. We apply the selection algorithm with constraints and cutoffs of $k = 5$ and $\tau < 0$ ($n = 6$), as well as an unfiltered collection ($n = 10$) (Table~\ref{tab:geographically-diverse-datasets-all}). While this offers greater geographic diversity than is typical in ML fairness benchmarks, it remains insufficient. As prior work has emphasized \citep{septiandri2023weird,mihalcea2025ai}, future data efforts must expand beyond WEIRD contexts while carefully balancing this goal with  ethical data practices.

\begin{figure}
    \centering
    \includegraphics[width=1.0\linewidth]{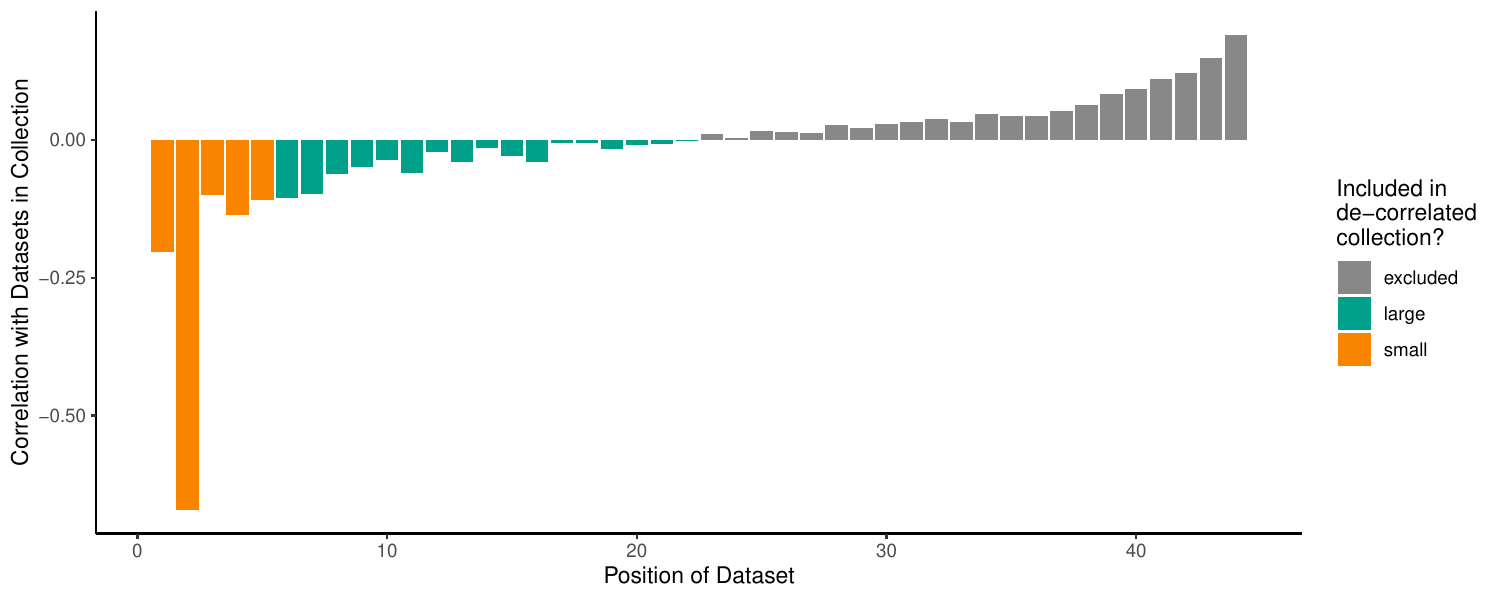}
    \caption{\textbf{A large number of negatively intercorrelated datasets is available for collection creation.} Average Spearman correlation of delta scores between the scenarios already in the collection and candidate scenarios at the time they are added to the collection. The very first scenario minimizes the average correlation with all other scenarios.}
    \label{fig:fig-bench-cors-1}
\end{figure}

\section{Limitations}
\label{sec:limitations}

While this work takes a substantial step toward improving reproducibility and empirical rigor in fair ML, it also operates within known constraints. Benchmarking, particularly in fairness research, can risk oversimplifying complex sociotechnical issues. Fairness cannot be fully captured by metrics or solved solely through optimization, and responsible development and evaluation of fair ML requires critical engagement with the broader context.

Our preprocessing and annotation decisions are not intended as universally optimal; their suitability depends on the specific dataset and use case. The experimental results presented here are illustrative rather than prescriptive—they demonstrate the kinds of analyses our corpus enables but are not meant to be definitive benchmarks.

Importantly, our dataset corpus is designed to be dynamic. Gaps in representation, especially with respect to geographic and demographic diversity, remain. We explicitly encourage community contributions of new datasets to help close these gaps (cf. Section~\ref{sec:corpus}). To support this, we provide a modular, versioned Python package that ensures transparency and reproducibility as the corpus evolves.

While our current focus is on tabular classification--a core setting in fair ML research--our framework is general. LLM evaluations, for example, may also benefit from the current tabular corpus through approaches such as folktexts \citep{cruz2025folktext}.
In future work, we aim to extend our methodology and infrastructure to other data modalities, including text and image domains.

\section{Discussion}
\label{sec:discussion}

We introduce \textit{FairGround}, a comprehensive framework, dataset corpus, and Python package developed to address long-standing challenges in fair ML research. By curating a diverse collection of 44 tabular datasets, encompassing 136 scenarios, and providing fairness-relevant metadata and reproducible preprocessing tools, FairGround enables transparent, rigorous, and extensible experimentation. The accompanying Python package supports reproducibility by exposing (and providing defaults for) key data processing decisions. We demonstrate its utility through a large-scale case study, illustrating how the framework facilitates robust comparative evaluations of debiasing techniques. Specifically, we show how our provided data collections better reflect the diverse performance of debiasing algorithms in comparison to collections currently used in fair ML research, while enabling new fairness analyses by connecting algorithm performance to dataset characteristics.

The significance of this work extends beyond its immediate technical contributions. By foregrounding the role of data infrastructure, FairGround highlights how dataset design, composition, and documentation fundamentally shape research trajectories and outcomes. These elements influence algorithmic behavior, reproducibility, and downstream system impact--making them critical to both scientific rigor and ethical responsibility.

Our framework is designed not only to support method development but also to position datasets as first-class research objects. It prompts researchers to interrogate representational biases, data provenance, and the implications of dataset selection--core concerns for equitable and socially responsible AI. In doing so, FairGround fosters deeper engagement with the sociotechnical dimensions of ML, encouraging reflection on how benchmarks reflect and reinforce power structures.

Additionally, FairGround lays essential groundwork for linking dataset characteristics to model fairness outcomes. This connection has important implications for anti-discrimination policy and regulation. For instance, under the EU AI Act, high-risk AI systems are subject to strict data governance requirements, including the obligation to assess datasets for bias and representational gaps \citep{eu2024aiact}. The metadata and fairness-relevant characteristics computed within FairGround can serve as a foundation for quantitative dataset documentation aligned with these legal mandates.

\impact{
%Authors must include a Broader Impact Statement, which should provide a concise, tangible portrayal of both the potential positive and negative societal consequences of their work. We refer to the submission guidelines for further details.

Our work aims to improve data practices in the field of algorithmic fairness, which in turn can lead to more robust and reproducible research, better and more ethical handling of datasets and increased transparency around dataset usage. By highlighting and quantifying the lack of geographic representation in popular datasets, we hope our work inspires the collection of novel and geographically diverse datasets. These positive changes have the possibility of affecting practices beyond research, ideally leading to the deployment of better and fairer algorithmic decision making and ML systems in production settings. Beyond the field of algorithmic fairness the \textit{FairGround} framework provides a template for other fields to start developing dataset corpora and collections.

While this work encourages better data practices, there is a risk of it contributing to a benchmarking culture overly focused on quantitative and superficial notions of fairness, which we explicitly want to warn against. While it is important to use a diverse collection of datasets for evaluation, it is equally important, especially in applied contexts, to be aware of the sociotechnical context (ML) systems are developed and deployed in.
}

% Acknowledgements and Disclosure of Funding should go at the end, before appendices and references
% All acknowledgements go at the end of the paper before appendices and references.
%Moreover, you are required to declare funding (financial activities supporting the
%submitted work) and competing interests (related financial activities outside the submitted work).
%More information about this disclosure can be found on the DMLR website.

\acks{We would like to thank F. Weber and A. Szimmat for their help in the annotation process.
The authors gratefully acknowledge the computational and data resources provided by the Leibniz Supercomputing Centre (www.lrz.de).

This work is supported by the DAAD programme Konrad Zuse Schools of Excellence in Artificial Intelligence, sponsored by the Federal Ministry of Education and Research and the Munich Center for Machine Learning (MCML).

This work is supported by BERD@NFDI and the Simons Institute for the Theory of Computing at the University of California, Berkeley.
}

\vskip 0.2in

\bibliography{bibliography}

%%%%%%%%%%%%%%%%%%%%%%%%%%%%%%%%%%%%%%%%%%%%%%%%%%%%%%%%%%%%

\newpage

\appendix

% Use letters in fig / table numbers
\counterwithin{figure}{section}
\counterwithin{table}{section}
\renewcommand\thefigure{\thesection\arabic{figure}}
\renewcommand\thetable{\thesection\arabic{table}}

%\section*{Technical Appendices and Supplementary Material}
%Technical appendices with additional results, figures, graphs and proofs may be submitted with the paper submission before the full submission deadline (see above), or as a separate PDF in the ZIP file below before the supplementary material deadline. There is no page limit for the technical appendices.

\section{Supplementary Tables}
\label{app:extra-tables}

This section includes supplementary tables that provide additional information supporting the results presented in the main text.

\begin{ThreePartTable}
\centering
\label{tab:datasets}

\newcolumntype{R}[1]{>{\raggedright\arraybackslash}p{#1}}

\begin{xltabular}{\textwidth}{r R{7cm} r r l}

\caption{Overview of datasets in the corpus. Row and column counts apply to the prepared data prior to further transformations.}\\
\toprule
  & Name and Citation & Rows & Columns & License\\
\midrule
\endfirsthead

%\caption[]{Overview of datasets in the corpus. Row and column counts apply to the prepared data prior to further transformations (cont.).}\\
\toprule
  & Name and Citation & Rows & Columns & License\\
\midrule
\endhead

\bottomrule
\endfoot

1 & Adult \citep{kohavi1996scaling} & 32,560 & 15 & CC BY 4.0\\
2 & Arrhythmia \citep{guvenir1998arrhythmia} & 451 & 280 & CC BY 4.0\\
3 & Bank (additional + full) \citep{moro2014data} & 41,188 & 21 & CC BY 4.0\\
4 & Bank (additional) \citep{moro2014data} & 4,119 & 21 & CC BY 4.0\\
5 & Bank (full) \citep{moro2014data} & 45,211 & 17 & CC BY 4.0\\
6 & Bank \citep{moro2014data} & 4,521 & 17 & CC BY 4.0\\
7 & Communities \citep{redmond2009communities} & 1,993 & 128 & CC BY 4.0\\
8 & Communities (unnormalized) \citep{lahoti2019operationalizing} & 2,214 & 147 & CC BY 4.0\\
9 & COMPAS (2 years) \citep{angwin2016machine} & 6,172 & 53 & ?\\
10 & COMPAS (2 years, violent) \citep{angwin2016machine} & 4,743 & 54 & ?\\
11 & COMPAS \citep{angwin2016machine} & 11,757 & 47 & ?\\
12 & CreditCard \citep{yeh2009default} & 30,000 & 25 & CC BY 4.0\\
13 & Drug \citep{fehrman2015drug} & 1,885 & 32 & CC BY 4.0\\
14 & Dutch \citep{le2022survey} & 60,420 & 12 & \tnote{a}\\
15 & German Credit \citep{hofmann1994statlog} & 1,000 & 21 & CC BY 4.0\\
16 & German Credit (numeric) \citep{hofmann1994statlog} & 1,000 & 25 & CC BY 4.0\\
17 & South German Credit \citep{gromping2019south} & 1,000 & 21 & CC BY 4.0\\
18 & German Credit (onehot) \citep{hofmann1994statlog} & 1,000 & 65 & Apache License\\
19 & Heart Disease \citep{janosi1988heart} & 303 & 14 & CC BY 4.0\\
20 & HMDA \citep{CFPB2022hmda} & 2,000,000 & 19 & ?\\
21 & Law School (tensorflow) \citep{wightman1998lsac} & 22,407 & 39 & CC BY-SA 4.0\\
22 & Law School (LeQuy) \citep{wightman1998lsac,le2022survey} & 18,692 & 12 & CC BY-SA 4.0\\
23 & MEPS (Panel 19, FY2015) \citep{ahrq2018meps} & 15,830 & 1,831 & \tnote{b}\\
24 & MEPS (Panel 20, FY2015) \citep{ahrq2018meps} & 17,570 & 1,831 & \tnote{b}\\
25 & MEPS (Panel 21, FY2016) \citep{ahrq2018meps} & 15,675 & 1,941 & \tnote{b}\\
26 & Nursery \citep{rajkovic1989nursery} & 12,960 & 9 & CC BY 4.0\\
27 & ricci \citep{miao2010did} & 118 & 5 & ?\\
28 & Stop, Question and Frisk Data \citep{NYPD2012} & 8,947 & 83 & \tnote{c}\\
29 & Chicago Strategic Subject List \citep{chicago2020strategic} & 398,684 & 48 & NA\\
30 & Student \citep{cortez2008using} & 395 & 33 & CC BY 4.0\\
31 & Student (Language) \citep{cortez2008using} & 649 & 33 & CC BY 4.0\\
32 & generate\_synthetic\_data \citep{zafar2017fairness} & 2,000 & 4 & GPL-3.0\\
33 & Lipton synthetic hiring dataset \citep{lipton2018does} & 2,000 & 4 & CC 0\\
34 & synth \citep{donini2018empirical} & 6,400 & 4 & ?\\
35 & Folktables ACSIncome \citep{ding2021retiring} & 1,664,500 & 11 & CC 0\\
36 & Folktables ACSPublicCoverage \citep{ding2021retiring} & 1,138,289 & 20 & CC 0\\
37 & Folktables ACSMobility \citep{ding2021retiring} & 620,937 & 22 & CC 0\\
38 & Folktables ACSEmployment \citep{ding2021retiring} & 3,236,107 & 17 & CC 0\\
39 & Folktables ACSTravelTime \citep{ding2021retiring} & 1,466,648 & 17 & CC 0\\
40 & Folktables ACSIncome (small) \citep{ding2021retiring} & 245,673 & 11 & CC 0\\
41 & Folktables ACSPublicCoverage (small) \citep{ding2021retiring} & 174,178 & 20 & CC 0\\
42 & Folktables ACSMobility (small) \citep{ding2021retiring} & 98,081 & 22 & CC 0\\
43 & Folktables ACSEmployment (small) \citep{ding2021retiring} & 478,236 & 17 & CC 0\\
44 & Folktables ACSTravelTime (small) \citep{ding2021retiring} & 216,385 & 17 & CC 0\\

\end{xltabular}

\begin{tablenotes}
\item[a] Copyright 2001, Centraal Bureau voor de Statistiek (CBS) (Statistics Netherlands) and Minnesota Population Center.
\item[b] See \url{https://meps.ahrq.gov/data_stats/data_use.jsp}.
\item[c] “All rights reserved”, see \url{https://www.nyc.gov/home/terms-of-use.page}.
\end{tablenotes}

\end{ThreePartTable}

\begin{table}
\centering
\caption{Countries represented in fair ML data. Each count represents a dataset that includes data from the specified country. There is one dataset representing data from across the world and one representing data from Hungary, Switzerland and the United States.}
\label{tab:country-dataset-distribution}
\begin{tabular}{lrr}
\toprule
Country & Count & Percentage (\%) \\
\midrule
United States & 28 & 59.57 \\
Portugal & 6 & 12.77 \\
Germany & 4 & 8.51 \\
N/A & 3 & 6.38 \\
Hungary, Switzerland \& United States & 1 & 2.13 \\
\addlinespace
Netherlands & 1 & 2.13 \\
Slovenia & 1 & 2.13 \\
Taiwan & 1 & 2.13 \\
Turkey & 1 & 2.13 \\
World & 1 & 2.13 \\
\bottomrule
\end{tabular}
\end{table}

\begin{table}[htbp]
\centering
\caption{Quantitative comparison of datasets available in different fairness libraries. *FairGround allows for the input of any custom fairness methods by users.}
\label{tab:tab-quant-comparison}
\newcolumntype{R}[1]{>{\raggedright\arraybackslash}p{#1}}
\begin{tabular}{lp{3cm}cccc}
\hline
 & & \multicolumn{2}{c}{\textbf{Number of}} & \textbf{Meta-} & \textbf{Collec-} \\
\textbf{Library} & \textbf{Main Focus} & \textbf{Datasets} & \textbf{Methods} & \textbf{Features} & \textbf{tions} \\
\hline
ABCFair & methods, metrics & 7 (5) & 10 & \ding{55} & \ding{55} \\
Aequitas Flow & methods, metrics, guides & 11 (11) & 10 & \ding{55} & \ding{55} \\
AIF360 & methods, metrics & 8 (8) & 15 & \ding{55} & \ding{55} \\
Fairlearn & methods, metrics, guides & 6 (4) & 6 & \ding{55} & \ding{55} \\
FairGround (ours) & data & 44 & 7* & \ding{51} & \ding{51} \\
\hline
\end{tabular}
\end{table}

\begin{table}[htbp]
\centering
\caption{Comparison of dataset collections in FairGround and other work, showing whether a debiasing method is ever the best performing method for any of the datasets for Equalized Odds Difference (left) and Demographic Parity Difference (right). For outside collections the closest matching scenarios within FairGround are selected.}
\label{tab:tab-copmarison-method-performance}
\small
\begin{threeparttable}
\begin{tabularx}{\textwidth}{l|*{4}{>{\centering\arraybackslash}X}|*{4}{>{\centering\arraybackslash}X}}
\hline
& \multicolumn{4}{c|}{\textbf{FairGround}} & & & & \\
& \footnotesize{\textbf{All}} & \footnotesize{\textbf{Open (all)}} & \footnotesize{\textbf{Open (lg.)}} & \footnotesize{\textbf{Open (sm.)}} & \footnotesize{\textbf{ABC Fair\textsuperscript{a}}} & \textbf{AIF 360\textsuperscript{b}} & \footnotesize{\textbf{Fried-ler}\textsuperscript{c}} & \footnotesize{\textbf{Typ. 3\textsuperscript{d}}} \\
\hline
\footnotesize{DisparateImpactRemover (pre)} & \cmark / \xmark & \cmark / \cmark & \cmark / \cmark & \cmark / \cmark & \cmark / \cmark & \xmark / \cmark & \cmark / \cmark & \xmark / \xmark \\
\footnotesize{LFR (pre)} & \cmark / \cmark & \cmark / \cmark & \cmark / \cmark & \cmark / \cmark & \xmark / \xmark & \cmark / \cmark & \cmark / \cmark & \cmark / \cmark \\
\footnotesize{GridSearchReduction (in)} & \cmark / \cmark & \cmark / \cmark & \cmark / \cmark & \cmark / \cmark & \cmark / \cmark & \cmark / \cmark & \cmark / \cmark & \cmark / \cmark \\
\footnotesize{AdversarialDebiasing (in)} & \cmark / \cmark & \cmark / \cmark & \cmark / \cmark & \cmark / \cmark & \cmark / \cmark & \cmark / \cmark & \cmark / \cmark & \cmark / \cmark \\
\footnotesize{MetaFairClassifier (in)} & \cmark / \cmark & \cmark / \cmark & \cmark / \cmark & \xmark / \cmark & \cmark / \cmark & \cmark / \cmark & \cmark / \cmark & \cmark / \cmark \\
\footnotesize{GerryFairClassifier (in)} & \cmark / \cmark & \cmark / \cmark & \cmark / \cmark & \cmark / \cmark & \cmark / \cmark & \cmark / \cmark & \xmark / \xmark & \xmark / \xmark \\
\footnotesize{CalibratedEqOdds (post)} & \cmark / \cmark & \cmark / \cmark & \cmark / \cmark & \cmark / \cmark & \cmark / \cmark & \xmark / \xmark & \xmark / \xmark & \xmark / \xmark \\
\hline
\footnotesize{\textbf{No. of Datasets}} & 44 & 32 & 16 & 5 & 5 & 7 & 5 & 3 \\
\hline
\end{tabularx}
\begin{tablenotes}
\small
\item[a] Five out of seven datasets in ABCFair \citep{defrance2024abcfair} are used.
\item[b] Seven out of eight datasets in AIF360 \citep{bellamy2018aif360} are used, the skipped dataset is available in FairGround, but used as a regression dataset in AIF360.
\item[c] \citet{friedler2019comparative}
\item[d] ``Typical 3'' refers to Adult, Compas and German Credit, the three most commonly used datasets in fairness research \citep{fabris2022algorithmic}.
\end{tablenotes}
\end{threeparttable}
\end{table}

% ==== Collection Tables ====

\begin{table}

\caption{\label{tab:de-correlated-datasets-large} Scenarios in the \textit{De-Correlated Datasets} collection. Column \textit{C} denotes collection membership: $k$ corresponds to the small collection with a cutoff value of $k = 5$; $\tau$ corresponds to the bigger collection with a cutoff value of $\tau = 0$. The larger collection encompasses the smaller one. Scenarios are listed based on insertion order.}
\centering
\begin{tabular}[t]{rll>{\raggedright\arraybackslash}p{3.8cm}l}
\toprule
  & C & Dataset & Sens. Attributes & Domain\\
\midrule
1 & $k$ & folktables\_acspubliccoverage & RAC1P & economics\\
2 & $k$ & heart\_disease & sex & cardiology\\
3 & $k$ & hmda & applicant\_sex\_name; applicant\_race\_name\_1 & finance\\
4 & $k$ & stop\_question\_and\_frisk\_data & SUSPECT\_SEX; SUSPECT\_RACE\_DESCRIPTION;  SUSPECT\_REPORTED\_AGE & law\\
5 & $k$ & folktables\_acsemployment\_small & RAC1P & economics\\
6 & $\tau$ & folktables\_acstraveltime & RAC1P & economics\\
7 & $\tau$ & compas & sex; age & law\\
8 & $\tau$ & folktables\_acsincome\_small & RAC1P & economics\\
9 & $\tau$ & compas\_2\_years & age & law\\
10 & $\tau$ & communities\_unnormalized & pct12-21 & law\\
11 & $\tau$ & arrhythmia & sex & cardiology\\
12 & $\tau$ & folktables\_acspubliccoverage\_small & RAC1P & economics\\
13 & $\tau$ & compas\_2\_years\_violent & age & law\\
14 & $\tau$ & south\_german\_credit & age; foreign\_worker & finance\\
15 & $\tau$ & dutch & age & demography\\
16 & $\tau$ & folktables\_acsmobility\_small & RAC1P & economics\\
17 & $\tau$ & law\_school\_tensorflow & gender & education\\
18 & $\tau$ & german\_credit\_onehot & <= 25 years & finance\\
19 & $\tau$ & communities & racePctAsian & law\\
20 & $\tau$ & nursery & finance & education\\
21 & $\tau$ & german\_credit\_numeric & age & finance\\
22 & $\tau$ & chicago\_strategic\_subject\_list & RACE CODE CD & law\\
\bottomrule
\end{tabular}
\end{table}

\begin{table}

\caption{\label{tab:permissively-licensed-datasets-all}Scenarios in the  \textit{Permissively Licensed Datasets} collection. Column \textit{C} denotes collection membership: $k$ corresponds to the small collection with a cutoff value of $k = 5$; $\tau$ corresponds to the bigger collection with a cutoff value of $\tau = 0$; an empty value corresponds to the full collection. The larger collections encompass the smaller ones. Scenarios are ordered based on when they were added to the collection.
}
\centering
\begin{tabular}[t]{rll>{\raggedright\arraybackslash}p{3.8cm}l}
\toprule
  & C & Dataset & Sens. Attributes & license\\
\midrule
1 & $k$ & folktables\_acspubliccoverage & RAC1P & CC 0\\
2 & $k$ & heart\_disease & sex & CC BY 4.0\\
3 & $k$ & communities\_unnormalized & pct12-21 & CC BY 4.0\\
4 & $k$ & lipton\_synthetic\_hiring\_dataset & sex & CC 0\\
5 & $k$ & bank & age; marital & CC BY 4.0\\
6 & $\tau$ & german\_credit\_onehot & > 25 years & Apache License\\
7 & $\tau$ & folktables\_acsincome & RAC1P & CC 0\\
8 & $\tau$ & south\_german\_credit & age & CC BY 4.0\\
9 & $\tau$ & folktables\_acsemployment\_small & RAC1P & CC 0\\
10 & $\tau$ & german\_credit\_numeric & age & CC BY 4.0\\
11 & $\tau$ & student & sex; age & CC BY 4.0\\
12 & $\tau$ & folktables\_acstraveltime\_small & RAC1P & CC 0\\
13 & $\tau$ & folktables\_acspubliccoverage\_small & RAC1P & CC 0\\
14 & $\tau$ & communities & agePct16t24 & CC BY 4.0\\
15 & $\tau$ & folktables\_acsmobility & RAC1P & CC 0\\
16 & $\tau$ & law\_school\_tensorflow & gender & CC BY-SA 4.0\\
17 &  & arrhythmia & sex & CC BY 4.0\\
18 &  & adult & race & CC BY 4.0\\
19 &  & nursery & finance; parents & CC BY 4.0\\
20 &  & folktables\_acsincome\_small & RAC1P & CC 0\\
21 &  & creditcard & SEX & CC BY 4.0\\
22 &  & folktables\_acsmobility\_small & RAC1P & CC 0\\
23 &  & student\_language & age & CC BY 4.0\\
24 &  & drug & ethnicity & CC BY 4.0\\
25 &  & law\_school\_lequy & racetxt; male & CC BY-SA 4.0\\
26 &  & folktables\_acstraveltime & RAC1P & CC 0\\
27 &  & bank\_additional\_full & age; marital & CC BY 4.0\\
28 &  & german\_credit & foreign\_worker & CC BY 4.0\\
29 &  & generate\_synthetic\_data & s1 & GPL-3.0\\
30 &  & bank\_additional & age & CC BY 4.0\\
31 &  & folktables\_acsemployment & RAC1P & CC 0\\
32 &  & bank\_full & age & CC BY 4.0\\
\bottomrule
\end{tabular}
\end{table}

\begin{table}

\caption{\label{tab:geographically-diverse-datasets-all}Scenarios in the  \textit{Geographically Diverse Datasets} collection. Column \textit{C} denotes collection membership: $k$ corresponds to the small collection with a cutoff value of $k = 5$; $\tau$ corresponds to the bigger collection with a cutoff value of $\tau = 0$; an empty value corresponds to the full collection. The larger collections encompass the smaller ones. Scenarios are ordered based on when they were added to the collection.}
\centering
\begin{tabular}[t]{rll>{\raggedright\arraybackslash}p{3.8cm}l}
\toprule
  & C & Dataset & Sens. Attributes & country\\
\midrule
1 & $k$ & folktables\_acspubliccoverage & RAC1P & USA\\
2 & $k$ & heart\_disease & sex & HUN;CHE;USA\\
3 & $k$ & dutch & age; citizenship & NLD\\
4 & $k$ & creditcard & SEX & TWN\\
5 & $k$ & german\_credit\_onehot & > 25 years & DEU\\
6 & $\tau$ & student & sex & PRT\\
7 &  & arrhythmia & sex & TUR\\
8 &  & nursery & finance; parents & SVN\\
9 &  & synth & sensible\_feature & NA\\
10 &  & drug & ethnicity & WORLD\\
\bottomrule
\end{tabular}
\end{table}

% ==== End of Collection Tables ====

% Clear all floats
\afterpage{\clearpage}

\section{Supplementary Figures}
\label{app:extra-figs}

This section contains supplementary figures that complement the primary results and provide further context for the analyses discussed in the main manuscript.

\begin{figure}[H]
    \centering
    \includegraphics[width=1.0\linewidth]{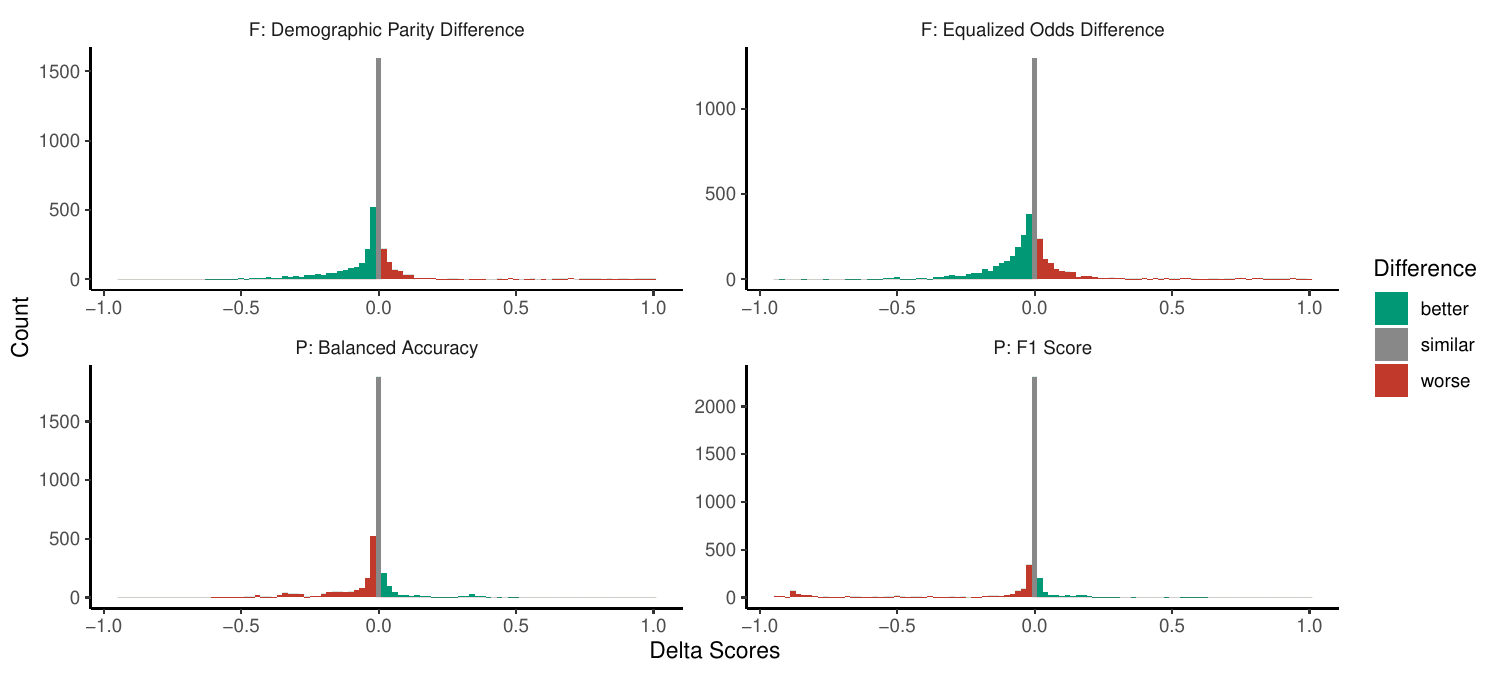}
    \caption{Delta scores across all four metrics are highly variable. Distribution of delta values for metrics of performance and fairness across different processing algorithms. Color-coding indicates whether the change is sizable (above an absolute threshold of 0.01) and corresponds to better (green) or worse (red) scores. For algorithmic fairness metrics lower scores are more desirable, whereas for metrics of performance higher scores are more desirable.}
    \label{fig:fig-hist-deltas-overview-1}
\end{figure}

\begin{figure}[H]
    \centering
    \includegraphics[width=\linewidth]{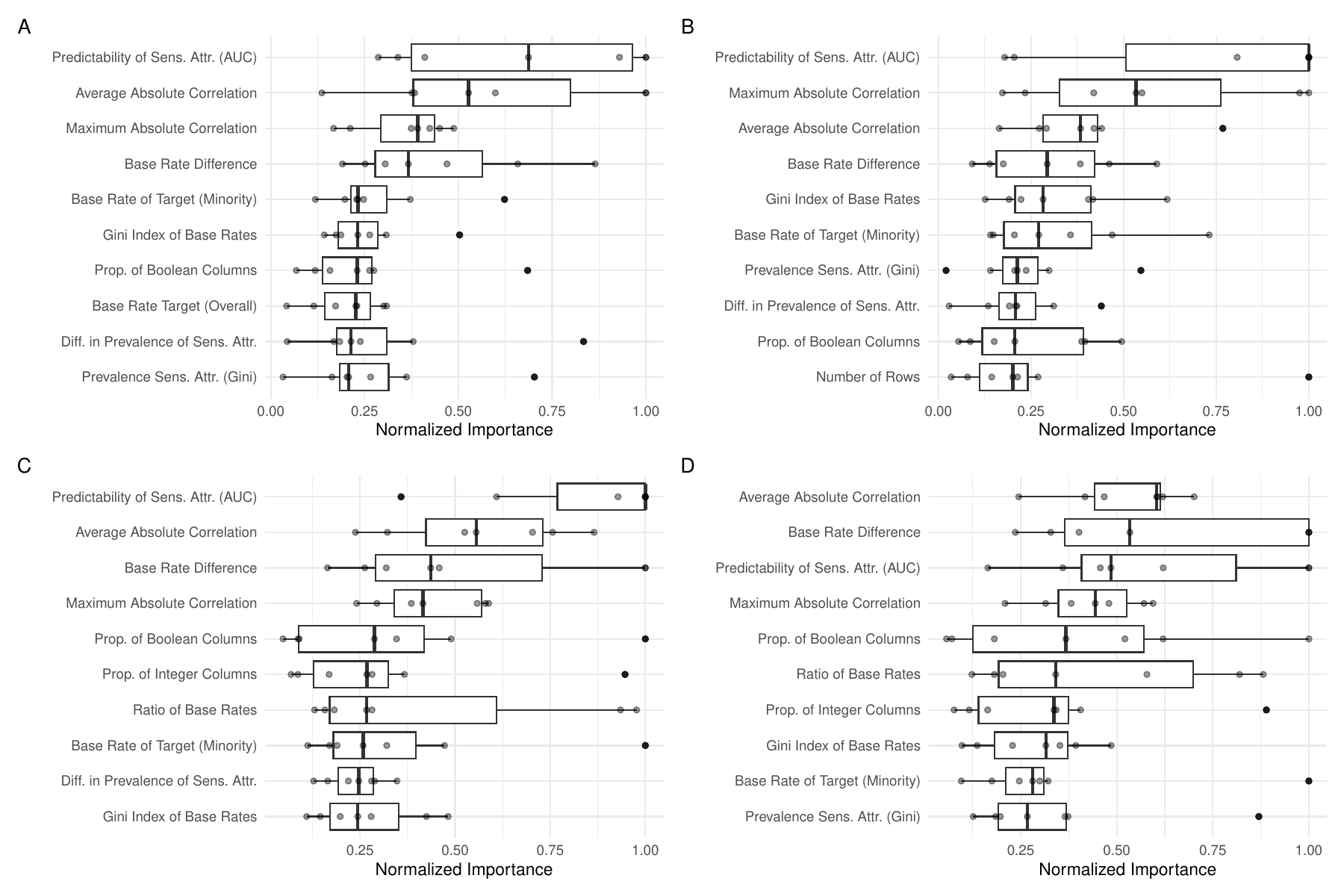}
    \caption{Normalized feature importance of the 10 most important computed metadata features to predict the difference in \textit{Balanced Accuracy} (A), \textit{F1 Score} (B), \textit{Equalized Odds Difference} (C) and \textit{Demographic Parity Difference} (D) across different processing methods.}
    \label{fig:fig-pred-varimp-combined-1}
\end{figure}

% Optional: Use [H] to force figures to stay in place.
\begin{figure}[H]
    \centering
    \includegraphics[width=\linewidth]{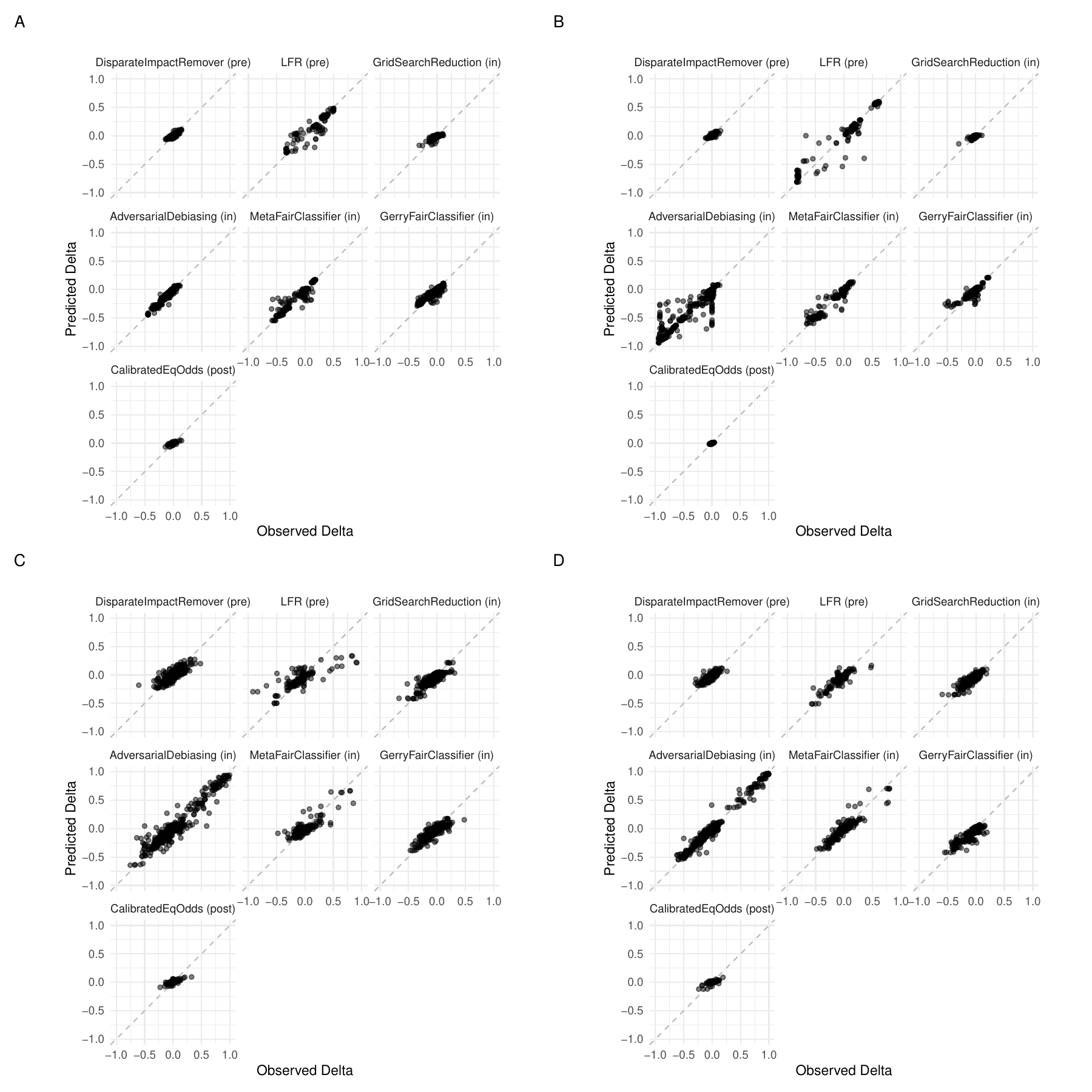}
    \caption{Comparison between observed and model-predicted values for \textit{Balanced Accuracy} (A), \textit{F1 Score} (B), \textit{Equalized Odds Difference} (C) and \textit{Demographic Parity Difference} (D) across different processing methods.}
    \label{fig:fig-pred-vs-observed-combined-1}
\end{figure}

\section{Technical Appendix}

\subsection{Metrics}

\begin{align}
    \text{Precision} &= \Pr(y=1|\hat{y}=1) \nonumber \\
    \text{Recall} &= \Pr(\hat{y}=1|y=1) \nonumber \\
    \text{Specificity} &= \Pr(\hat{y}=0|y=0) \nonumber
\end{align}

We use Balanced Accuracy (bAcc; Eq. \ref{eq:bacc}) and F1 Score (Eq. \ref{eq:f1}) as measures of performance. The two performance metrics are defined as follows:

\begin{align}
    \text{bACC} &= \frac{\text{Specificity} + \text{Recall}}{2}  \label{eq:bacc} \\
    \text{F1 Score} &= \frac{2}{\text{Precision}^{-1}+\text{Recall}^{-1}} \label{eq:f1}
\end{align}

We use Equalized Odds Difference (EOD; Eq. \ref{eq:eod}) and Demographic Parity Difference (DPD; Eq. \ref{eq:dpd}) as measures of algorithmic fairness. The two fairness metrics are defined as follows:

\begin{align}
    \text{EOD} &= \max_g \Pr(\hat{y}=1|y=1, S=g) - \min_g \Pr(\hat{y}=1|y=1, S=g)\label{eq:eod} \\
    \text{DPD} &= \max_g \Pr(\hat{y}=1|S=g) - \min_g \Pr(\hat{y}=1|S=g)\label{eq:dpd}
\end{align}

When comparing different fairness aware methods, we use delta scores ($\Delta_{a,b}$) for their comparison. These scores are computed for each performance and fairness metric and are defined as follows:

\begin{equation}
\label{eq:delta_score}
\Delta_{a,b} = score_{a,b} - score_{a,\text{baseline}}
\end{equation}

\subsection{Selection Algorithm}
\label{sec:selection-algorithm}

Given the corpus of datasets and their associated scenarios $\mathcal{D} = \{D_1, D_2, \dots, D_N\}$, where each dataset $D_i$ consists of a set of scenarios $D_i = \{s_{i1}, s_{i2}, \dots\}$, the goal is to construct a collection of scenarios $\mathcal{C}$ such that the pairwise spearman correlations of delta scores (Eq.~\ref{eq:delta_score}), across \textit{Balanced Accuracy} (Eq.~\ref{eq:bacc}), \textit{F1 Score} (Eq.~\ref{eq:f1}), \textit{Equalized Odds Difference} (Eq.~\ref{eq:eod}) and \textit{Demographic Parity Difference} (Eq.~\ref{eq:dpd}) between members of $\mathcal{C}$ are as low as possible across different families of fair ML algorithms (\textit{Learning Fair Representations} \citep{zemel2013learning}, \textit{Disparate Impact Remover} \citep{feldman2015certifying}, \textit{Adversarial Debiasing} \citep{zhang2018mitigating}, \textit{Meta-Algorithm} \citep{celis2019classification}, \textit{Rich Subgroup Fairness / GerryFair} \citep{kearns2018preventing}, \textit{Grid Search Reduction} \citep{agarwal2018reductions}, \textit{Group-Specific Thresholds} \citep{hardt2016equality}). To control the number of scenarios in $\mathcal{C}$, we use either a fixed number $k$ or a correlation threshold $\tau$. While this work uses only one of these constraints at a time, they can be combined if desired. The algorithm proceeds as follows:

\begin{enumerate}
    \item Let $r_{ab}$ denote the \textit{Spearman rank correlation} between scenarios $s_a$ and $s_b$, where $s_a, s_b \in \bigcup_{i=1}^N D_i$.

    \item For each scenario $s_a$, compute the average Spearman correlation to all other scenarios:
    \[
    \bar{r}_a = \frac{1}{M - 1} \sum_{b \neq a} r_{ab}
    \]
    where $M$ is the total number of scenarios in the corpus. Select the scenario $s_m$ with the lowest average correlation:
    \[
    m = \arg\min_a \bar{r}_a
    \]
    Initialize the selected set $\mathcal{C} = \{s_m\}$, and the remaining pool $\mathcal{R} = \left(\bigcup_{i=1}^N D_i\right) \setminus D_{i(m)}$, where $D_{i(m)}$ is the dataset containing scenario $s_m$.

    \item Repeat the following until $|\mathcal{C}| = k$ or no candidate in $\mathcal{R}$ has an average Spearman correlation strictly less than $\tau$ with all members of $\mathcal{C}$:
    \begin{enumerate}
        \item For each scenario $s_j \in \mathcal{R}$, compute the average correlation with the current set $\mathcal{C}$:
        \[
        \bar{r}_{j\mathcal{C}} = \frac{1}{|\mathcal{C}|} \sum_{s_i \in \mathcal{C}} r_{ij}
        \]
        \item Identify the scenario $s_{j^*}$ with the lowest such average:
        \[
        j^* = \arg\min_{j \in \mathcal{R}} \bar{r}_{j\mathcal{C}}
        \]
        \item If $\bar{r}_{j^*\mathcal{C}} < \tau$, add $s_{j^*}$ to $\mathcal{C}$, and remove all scenarios in the same dataset $D_{i(j^*)}$ from $\mathcal{R}$.
    \end{enumerate}

    \item The algorithm terminates when $|\mathcal{C}| = k$ or no remaining scenario has an average Spearman correlation below $\tau$ with the current set $\mathcal{C}$. The resulting subset $\mathcal{C}$ is returned as the final collection of minimally correlated scenarios.
\end{enumerate}

\subsection{Example Code using the Package}
\label{sec:package-examples}

The following subsection contains exemplary code illustrating the usage of the Python package. We recommend readers to review the online package documentation at \url{https://brave-ocean-078c2100f.6.azurestaticapps.net/} for a more in-depth description of the package's functions.

\subsubsection{Using a Dataset}
%\begin{minted}{python}
\begin{verbatim}

from fairml_datasets import Dataset

# Get the dataset
dataset = Dataset.from_id("folktables_acsemployment")

# Load as pandas DataFrame
df = dataset.load()  # or df = dataset.to_pandas()
print(f"Dataset shape: {df.shape}")

# Get the target column
target_column = dataset.get_target_column()
print(f"Target column: {target_column}")

# Get sensitive attributes (before transformation)
sensitive_columns_org = dataset.sensitive_columns

# Transform to e.g. impute missing data
df_transformed, transformation_info = dataset.transform(df)
# Sensitive columns may change due to transformation
sensitive_columns = transformation_info.sensitive_columns

# Split into train and test sets
train_df, test_df = dataset.train_test_split(df, test_size=0.3)

# Run analyses on the data
\end{verbatim}
%\end{minted}

\subsubsection{Using a Collection of Datasets / Scenarios}

%\begin{minted}{python}
\begin{verbatim}

from fairml_datasets.collections import DeCorrelatedSmall

collection = DeCorrelatedSmall()

# The collection consists of scenarios
for scenario in collection:
    # Each scenario behaves just like a dataset

    # Load as pandas DataFrame
    df = scenario.load()  # or df = scenario.to_pandas()
    print(f"Dataset shape: {df.shape}")

    # Get the target column
    target_column = scenario.get_target_column()
    print(f"Target column: {target_column}")

    # Get sensitive attributes (before transformation)
    sensitive_columns_org = scenario.sensitive_columns

    # Transform to  e.g. impute missing data
    df_transformed, transformation_info = scenario.transform(df)
    # Sensitive columns may change due to transformation
    sensitive_columns = transformation_info['sensitive_columns']

    # Split into train and test sets
    train_df, test_df = scenario.train_test_split(df, test_size=0.3)
    
    # Run analyses on the data
\end{verbatim}
%\end{minted}

\subsection{Annotation Procedure}
\label{sec:annotation-procedure}

We started the annotation process by collecting all tabular datasets used for fair classification tasks in a large survey of fair ML datasets \citep{fabris2022algorithmic}. This provided a list of $n = 37$ unique datasets. Additionally, we added the folktables \citep{ding2021retiring} collection of datasets, due to its recent popularity and as the datasets specifically try to address issues in the most popular dataset in the survey: \textit{Adult} \citep{kohavi1996scaling}.

For each dataset, we annotated the information required to practically use the dataset in a fair classification task, as well as key qualitative and quantitative data regarding the information represented in each dataset. During this process, a critical issue quickly became apparent: While datasets are commonly referenced by name as if they were uniquely identified, this is often not the case in practice. A striking example is the widely used Bank dataset, one of the most frequently cited datasets in Fair ML \citep{fabris2022algorithmic}. Although typically referred to as Bank or Bank Marketing, the primary source\footnote{\url{https://archive.ics.uci.edu/dataset/222/bank+marketing}} actually comprises four distinct datasets, each differing in their respective number of instances and attributes. Recognizing this ambiguity, we adapted our annotation methodology to explicitly capture dataset variants, significantly increasing the number of distinct datasets in the corpus. In our framework, we treat these variants as separate datasets while preserving their connection to maintain clarity and traceability.

When collecting the information required  to download and load datasets, we were forced to exclude $n = 11$ datasets due to data not being publicly available or with restricted access. We excluded a further $n = 18$ datasets, if there were issues with recreating how a dataset was generated or the dataset's usage did not fit into schema of a "classic" fairML classification task including features, a target column and sensitive attribute(s). A detailed breakdown of excluded datasets and the reasons for their exclusion is available in Section~\ref{sec:excluded-datasets}.

After exclusion of non-eligible datasets and inclusion of different variants, we arrive at a list of $N = 44$ datasets.

Datasets were annotated by two of the authors with help from research assistants. A random subset of annotations was reviewed by a third author.

\subsection{Annotated Columns}
\label{sec:annotation-columns-appendix}

The following section provides descriptions of columns which were manually annotated for each dataset in the corpus.

\textbf{new\_dataset\_id} \quad A unique identifier for each dataset. Usually derived from the dataset name.

\textbf{dataset\_name} \quad An official, common, or known name of the dataset that is unique across datasets.

\textbf{base\_dataset\_name} \quad In case there are different variants of the same dataset, this field holds a common name to group all these variants together.

\textbf{description\_public} \quad This is a free-text field reporting (1) the aim/purpose of a data artifact (i.e., why it was developed/collected), as stated by curators or inferred from context; (2) a high-level description of the available features; (3) the labeling procedure for annotated attributes, with special attention to sensitive ones, if any; (4) the envisioned ML task, if any.

\textbf{notes\_public} \quad Any notes or comments regarding this dataset / task combination.

\textbf{dataset\_aliases} \quad Any names that this dataset is called by. While 'dataset\_name' only contains the single most common name, this field holds possible aliases used to reference this dataset.

\textbf{affiliation} \quad Affiliation of the creators of the dataset. Based on reports, articles, or official web pages presenting the dataset.

\textbf{domain\_class\_main} \quad The main field where the data is used (e.g., computer vision for ImageNet) or the field studying the processes and phenomena that produced the dataset (e.g., radiology for CheXpert).

\textbf{domain\_class\_multi} \quad The primary fields where the data is used (e.g., computer vision for ImageNet) or the fields studying the processes and phenomena that produced the dataset (e.g., radiology for CheXpert). Multiple domains are possible in this feature.

\textbf{domain\_freetext} \quad Fine-grained domain of the prediction task. Summarized with 1 - 2 words.

\textbf{sample\_size} \quad Dataset cardinality. Rough estimate of the size of the dataset.

\textbf{year\_last\_updated} \quad The last known update to the dataset. For resources whose collection and curation are ongoing (e.g., HMDA), we write “present”.

\textbf{years\_data} \quad The timespan covered in the data. This refers to the "social realities" captured in the data i.e., data from which year(s) is present in the data.

\textbf{citation} \quad The main / official source to cite this dataset in BibTeX format. For synthetic datasets, this refers to the original paper where the dataset was first introduced.

\textbf{main\_url} \quad The main landing page or website related to the dataset. This is a website with information on the dataset and not the dataset itself, which is referenced via 'download\_url'.

\textbf{related\_urls} \quad List of related links and resources to the dataset.

\textbf{license} \quad Under which license is the dataset made available? A "?" indicates that no license was found.

\textbf{continent} \quad Continent(s) where the dataset is sourced. In two-letter format. If "n/a", this concept is not applicable for a dataset (e.g., a synthetic one).

\textbf{country} \quad Countrie(s) where the dataset is sourced. In ISO3 format. If "n/a", this concept is not applicable for a dataset (e.g., a synthetic one).

\textbf{dataset\_variant\_id} \quad This ID is used to identify different datasets belonging to the same original dataset e.g., COMPAS has 3 unique smaller datasets belonging to this one bigger one. In cases like this, each smaller dataset gets its own dataset\_variant\_id.

\textbf{dataset\_variant\_description} \quad Description outlining how this "sub-dataset" is different from the others. Only filled out if there are multiple "dataset\_variant\_ids".

\textbf{is\_accessible} \quad Is the dataset publicly accessible? "Manual download" indicates that an automated download is not possible.

\textbf{download\_url} \quad URL to the dataset file itself, if it is publicly accessible.

\textbf{custom\_download} \quad Are there some extra steps needed to download the dataset itself, e.g., unpacking a ZIP archive?

\textbf{filename\_raw} \quad Filename of the dataset for downloading it or finding it in a ZIP archive.

\textbf{format} \quad Format of the dataset. Corresponds to the format the data is in, not the extension of the dataset e.g., CSV for comma-separated-values, TSV for tab-separated-values, FIXED-WIDTH for fixed-width formats etc.

\textbf{colnames} \quad Column names to use if the dataset file does not include them.

\textbf{processing} \quad Does the dataset need some special pre-processing to be in the correct format?

\textbf{sensitive\_attributes} \quad Sensitive attributes that are \textit{available} in the dataset. Supports multiple entries, separated with a semicolon and a space: '; '.

\textbf{typical\_col\_sensitive} \quad All columns containing available sensitive attributes and the information they contain in a categorical fashion. Covering the attributes listed in 'sensitive\_attributes'. Formatted as a JSON dictionary.

\textbf{typical\_col\_features} \quad All columns typically used as features / predictors. Either a list of column names indicating a positive selection or a list of column names prefixed with a \texttt{-} indicating a negative selection i.e. all columns except the listed ones. A \texttt{-} indicates using all available columns (except the target).

\textbf{typical\_col\_target} \quad Column(s) which are being predicted. If more than one, separated by semicolons.

\textbf{target\_lvl\_good} \quad Which value of the target variable is considered desirable? Desirable here means good for any person impacted by a system built using this data.

\textbf{target\_lvl\_bad} \quad Which value of the target variable is considered undesirable? Undesirable here means bad for any person impacted by a system built using this data.

\textbf{dataset\_size} \quad Whether a dataset is exceptionally large.

\subsection{Computed Metadata}
\label{sec:computed-meta-data}

The following section provides descriptions of the computed metadata features which are implemented in the Python package and computed for each of the datasets in the corpus. The technical implementation can be reviewed in the publicly available source code of the package.

% Fix overvlows by reducing LaTeX typesetting strictness
\emergencystretch 10em

\textbf{Size}\quad As \cite{ding2021retiring} note, increasing dataset size does not necessarily reduce observed disparities due to persistent structural inequalities. We try to cover a broad range of dataset sizes in our corpus and compute dataset sizes by rows (samples) and columns (attributes) of both prepared (\texttt{meta\_pretrans\_n\_rows, meta\_pretrans\_n\_cols}) and transformed datasets (\texttt{meta\_n\_rows, meta\_n\_cols}).
% (meta_pretrans_n_rows, meta_pretrans_n_cols, meta_n_rows, meta_n_cols)

\textbf{Missing values}\quad To address potential bias from missing data \citep[e.g. see][]{pessach2022review, wang2021analyzing, martinez2019fairness}, we calculate the fraction of missing data per dataset. Metadata was computed prior to processing to assess the proportion of rows (\texttt{meta\_pretrans\_prop\_NA\_rows}), columns (\texttt{meta\_pretrans\_prop\_NA\_cols}) and cells (\texttt{meta\_pretrans\_prop\_NA\_cells}) that contain missing values. We further calculate missingness within each group of the protected attribute (only when binarizing; \texttt{meta\_prop\_NA\_sens\_minority, meta\_prop\_NA\_sens\_majority}).
% meta_pretrans_prop_NA_rows, meta_pretrans_prop_NA_cols, meta_pretrans_prop_NA_cells, meta_prop_NA_sens_majority, meta_prop_NA_sens_minority

\textbf{Attribute types}\quad We calculate the proportions of different numeric (\texttt{meta\_prop\_cols\_float, meta\_prop\_cols\_int}) and logical (\texttt{meta\_prop\_cols\_bool}) data types in the data to assess their potential influence. 
% meta_prop_cols_float, meta_prop_cols_int, meta_prop_cols_bool

\textbf{Sensitive AUC} \quad Non-sensitive attributes can act as proxies for sensitive ones \citep[e.g. see][]{pessach2022review, mehrabi2021survey, fawkes2024fragility}. Identifying and addressing such proxies can help mitigate unfairness \citep{pessach2022review, matloff2022novel}. To assess this, we define \textit{Sensitive AUC} as the ROC-AUC of a random forest model \citep{ho1995random} trained to predict the sensitive attribute using only non-sensitive features (\texttt{meta\_sens\_predictability\_roc\_auc}). A higher Sensitive AUC suggests that non-sensitive attributes may encode sensitive information.
% meta_sens_predictability_roc_auc

\textbf{Bivariate correlations}\quad Serving as an additional indicator of potential proxy variables, we computed the correlation between each non-sensitive feature and the sensitive attribute, using the average and maximum correlation values (\texttt{meta\_average\_absolute\_correlation, meta\_maximum\_absolute\_correlation}).
% meta_average_absolute_correlation, meta_maximum_absolute_correlation

\textbf{Number of protected groups}\quad Some fairness methods require binary representations of protected attributes, leading to the binarization of categorical or numerical sensitive attributes during preprocessing. Documenting the original number of protected groups before processing (\texttt{meta\_pretrans\_unique\_group\_counts\_pre\_agg}) helps track this process and may provide insight into how such simplifications affect the performance and suitability of fairness methods.
% meta_pretrans_unique_group_counts_pre_agg

\textbf{Prevalence}\quad We computed the proportions of minority and majority groups within the dataset (only when binarizing; \texttt{meta\_prev\_sens\_minority, meta\_prev\_sens\_majority}), along with the absolute difference between them (\texttt{meta\_prev\_sens\_difference}) and the imbalance ratio (\texttt{meta\_prev\_sens\_ratio}). A smaller absolute difference and an imbalance ratio closer to 1 indicate a more balanced distribution of the sensitive attribute.

\textbf{Base Rate}\quad Similar to prevalence, we computed the probability of the favorable outcome overall (\texttt{meta\_base\_rate\_target}) and for each group (only when binarizing; \\
\texttt{meta\_base\_rate\_target\_sens\_minority}, \texttt{meta\_base\_rate\_target\_sens\_majority}) along with the absolute difference (\texttt{meta\_base\_rate\_difference}) and ratio (\texttt{meta\_base\_rate\_ratio}) between them.

\textbf{Gini-Simpson Index}\quad The Gini-Simpson Index measures the probability that two randomly selected individuals belong to different groups. Similar indices have been previously used by \cite{mecati2023measuring} and \cite{vetro2021data} to assess balance and detect potential unfairness in datasets. We compute the Gini-Simpson Index for both group prevalence and base rates

\[
GS = 1 - \sum_i p_i^2,
\]

where $p_i$ is the proportion of instances in group $i \in \{1, 2\}$ (protected or non-protected). For prevalence, this is is the proportion of individuals per group relative to the entire dataset (\texttt{meta\_prev\_sens\_gini}). For base rates, $p_i$ denotes the proportion of favorable outcomes within each group (\texttt{meta\_base\_rate\_sens\_gini}).
% meta_prev_sens_gini, meta_base_rate_sens_gini

% Revert typesetting standard back to normal
\fussy

\subsection{Excluded Datasets}
\label{sec:excluded-datasets}

This subsection contains explanations for additional datasets that were excluded from the corpus. The annotation procedure is described in detail in Section~\ref{sec:annotation-procedure}.

\textbf{2016 Presidential Elections (2 datasets)} \quad This dataset from the FiveThirtyEight 2016 Election Forecast was developed with the goal of providing an aggregated estimate of the probability that Trump/Clinton wins the 2016 election based on multiple polls, weighting each input according to sample size, recency, and historical accuracy of the polling organization. For each poll, the dataset provides the period of data collection, its sample size, the pollster conducting it, their rating, and a url linking to the source data. The dataset does not contain any sensitive attributes and was therefore excluded. One annotated but excluded dataset came from ABC News, and another, potentially deviating, from \citep{sabato2020bounding}.
% 2016_presidential_elections version Sabato & Yom-Tov (Microsoft & BGU) and ABC news

\textbf{Cancer Cases and Deaths (3 datasets)} \quad The main dataset reports state-level cancer prevalence for 18 cancer types, based on data from the CDC’s NPCR and the NCI’s SEER program. Mortality data come from the CDC’s National Vital Statistics System. As it contains only aggregated data on state-level, it was excluded from our analysis. Two additional datasets provided the source data on new cases and deaths. As neither was used in isolation in our annotations, both were excluded with the main dataset.
% cancer, cancer_new_case_rates, cancer_new_death_rates

\textbf{Clinical Annotations / Warfarin Dosage / PharmaGKB (4 datasets)} \quad The data, collected by the International Warfarin Pharmacogenetics Consortium and co-curated by PharmGKB, was used to study algorithmic estimation of optimal warfarin dosage. The original data includes thousands of patient demographics, comorbidities, medications, genetics, and effective warfarin doses. However, the available datasets do not contain demographic details and only a specialty group column indicates few pediatric cases. Due to the absence of sensitive attributes, these datasets were excluded. The excluded datasets comprised: 1) meta-data for each clinical annotation; 2) genotype/allele-based annotation text with CPIC-assigned function, if available; 3) supporting annotation details (variant, guideline, label); and 4) clinical annotation history with creation and update dates. 
% clinical_annotations_warfarin_dosage, pharmgkb

\textbf{COMPAS (4 datasets)} \quad We retain the original COMPAS data published by ProPublica \cite{angwin2016machine}. Specific versions of the COMPAS dataset were excluded, including an unofficial version published on Kaggle, used in one reviewed study \citep{jabbari2020empirical}, and two others, each appearing in a single paper \citep{wang2019repairing, mandal2020ensuring}, due to a lack of clarity in the differences and processing from the original ProPublica release. The COMPAS repository\footnote{\url{https://github.com/propublica/compas-analysis}} also includes a file with "raw" scores, named \texttt{compas-scores-raw.csv}, which we decided not to include, as it is not further utilized in the analysis. % A raw version published by the authors alongside their main dataset was also excluded. % (e.g., differently formatted date variable, later removed middle name column)
% compas_kaggle, compas_train_test_val_csv, compas_proxy_data_csv, compas raw

\textbf{FICO Credit Score, Credit Score Performance (2 datasets)} \quad The dataset originates from a 2007 Federal Reserve report to the US Congress on credit scoring and its effects on the availability and affordability of credit. The collection, creation, processing, and aggregation was carried out by the working group and is based on a sample of 301,536 TransUnion TransRisk scores from 2003. The dataset contains only aggregated statistics per FICO score and race/ethnicity group and was therefore excluded. A second version with unclear differences was also excluded.
% creditscore, creditscore_performance

\textbf{Fifa 20 Complete Player} \quad This dataset was scraped by Stefano Leone and shared on Kaggle. It contains player data from FIFA Career Mode (FIFA 15-20). We excluded this dataset, because relevant sensitive attributes and target variables were unclear. A paper by \cite{awasthi2021evaluating} created a sensitive attribute by predicting nationality from player names using LSTM, an approach that could introduce unnecessary uncertainty and therefore may have reduced comparability.
% fifa_20_complete_player

\textbf{Pima Diabetes} \quad This dataset was derived from a medical study of Native Americans from the Gila River Community, often called Pima. Conducted by the National Institute of Diabetes and Digestive and Kidney Diseases since the 1960s, the study found a large prevalence of \textit{diabetes mellitus} in this population. The dataset includes a subset of the original study, focusing on women of age 21 or older. It reports diabetes test results and eight key risk factors, such as number of pregnancies, skin thickness, and BMI. Relevant sensitive attributes were not clear based on the papers we reviewed, so we decided to exclude the dataset.
% pima_diabetes

\textbf{US Census (1990)} \quad This dataset is derived from the 1990 US Census. In the reviewed literature, the classification task was often unclear or unsuitable for our analysis goals \citep[e.g.,][]{sabato2020bounding}. Another meta-analysis referenced 25 selected numeric attributes without specifying them.
% us_census_data_1990

\subsection{Computational Infrastructure}

\label{sec:exp-infrastructure}

% https://doku.lrz.de/1-general-description-and-resources-10746641.html

Experiments were run on a shared Linux compute cluster with partitions and compute infrastructure chosen based on availability of resources. Experiments were run as four consecutive jobs, the first running experiments at high concurrency and the later re-running errored out experiments at lower concurrency.

The first job was run on a node with access to 76 CPU cores and 512 GB of memory over a duration of 11 hours. Later jobs were run on a node with 96 CPUs and 1 TB of memory, using 5-fold parallelism and a maximum execution time of 2.5 hours for the second and third run and 5 hours for the last run. Experiments were conducted using only CPU compute.

\subsection{Software}

\label{sec:software}

Simulation experiments were conducted using Python \citep{python310} version 3.10 and the Python package \texttt{multiversum} \citep{simson2024multiversum} version 0.7.0. We used the implementations of fairness-aware processing methods from the package \texttt{AIF360} \citep{bellamy2018aif360} and used \texttt{scikit-learn} \citep{pedregosa2011scikit} to fit logistic regressions. Data were processed using the newly developed \texttt{fairml\_datasets} package, utilizing \texttt{pandas} \citep{mckinney2010data}, \texttt{fastparquet} \citep{durantfastparquet} and \texttt{scikit-learn}. Multiple other packages were utilized as (peer) dependencies of the named packages. We use \texttt{uv} \citep{marsh2024uv} for virtual environment management.

Results from the experiments were analysed using R version 4.4.1 \citep{r2024} with packages from the \texttt{tidyverse} \citep{wickham2019welcome}, \texttt{patchwork} \citep{pederson2024patchwork} and \texttt{tidymodels} \citep{kuhn2020tidymodels}. Color schemes are used from the R packages \texttt{awtools} \citep{wehrwein2025awtools} and \texttt{wesanderson} \citep{ram2023wesanderson}. We use \texttt{renv} for virtual environment management.

Lockfiles for both Python and R packages are provided with the codebase.

Experiments were executed using a docker container converted to the \texttt{enroot} format\footnote{https://github.com/NVIDIA/enroot}.

\end{document}